 \documentclass[final,5p,times,twocolumn]{elsarticle}

\usepackage{amssymb}
\usepackage{amsthm}
\usepackage{amsmath} 
\usepackage{booktabs} 
\usepackage{hyperref}
\usepackage{multirow}
\usepackage{graphicx}
\usepackage{todonotes}
\usepackage{xcolor}
\usepackage{color,soul}

\journal{Image and Vision Computing}

\begin{document}

\begin{frontmatter}


\author[label1]{Ziya Ata Yazıcı\corref{cor1}}
\ead{yaziciz21@itu.edu.tr}

\author[label1]{İlkay Öksüz}
\ead{oksuzilkay@itu.edu.tr}

\author[label1,label2]{Hazım Kemal Ekenel}
\ead{ekenel@itu.edu.tr}
\ead{hekenel@qu.edu.qa}
\cortext[cor1]{Corresponding Author}

\title{GLIMS: Attention-Guided Lightweight Multi-Scale Hybrid Network for\\Volumetric Semantic Segmentation}

\affiliation[label1]{organization={Istanbul Technical University},
            addressline={Department of Computer Engineering},
             city={Istanbul},
             postcode={34467},
             country={Turkey}}

\affiliation[label2]{organization={Qatar University},
            addressline={Department of Computer Science and Engineering},
             city={Doha},
             postcode={2713},
             country={Qatar}}

\begin{abstract}

Convolutional Neural Networks (CNNs) have become widely adopted for medical image segmentation tasks, demonstrating promising performance. However, the inherent inductive biases in convolutional architectures limit their ability to model long-range dependencies and spatial correlations. While recent transformer-based architectures address these limitations by leveraging self-attention mechanisms to encode long-range dependencies and learn expressive representations, they often struggle to extract low-level features and are highly dependent on data availability. This motivated us for the development of GLIMS, a data-efficient attention-guided hybrid volumetric segmentation network. GLIMS utilizes Dilated Feature Aggregator Convolutional Blocks (DACB) to capture local-global feature correlations efficiently. Furthermore, the incorporated Swin Transformer-based bottleneck bridges the local and global features to improve the robustness of the model. Additionally, GLIMS employs an attention-guided segmentation approach through Channel and Spatial-Wise Attention Blocks (CSAB) to localize expressive features for fine-grained border segmentation. Quantitative and qualitative results on glioblastoma and multi-organ CT segmentation tasks demonstrate GLIMS' effectiveness in terms of complexity and accuracy. GLIMS demonstrated outstanding performance on BraTS2021 and BTCV datasets, surpassing the performance of Swin UNETR. Notably, GLIMS achieved this high performance with a significantly reduced number of trainable parameters. Specifically, GLIMS has 47.16M trainable parameters and 72.30G FLOPs, while Swin UNETR has 61.98M trainable parameters and 394.84G FLOPs. The code is publicly available on \url{https://github.com/yaziciz/GLIMS}.

\end{abstract}









\end{frontmatter}


\section{Introduction}

Medical image segmentation is a highly-studied research area due to its pivotal role in computer-aided diagnosis \cite{wang2022medical, zhou2021review}. Medical images are instrumental in revealing morphological lesions and functional changes. Thus, the precise segmentation of such anatomical regions is critical for effective pre-treatment diagnosis, treatment planning, and post-treatment assessment by physicians.

The segmentation task involves the precise classification of interested regions in medical images, pixel-wise or voxel-wise.
Over the past decades, numerous efforts have been dedicated to developing efficient and robust segmentation techniques, yielding significant advancements. Leveraging their capacity to extract image features, Convolutional Neural Networks (CNN) have found extensive application in diverse image segmentation tasks \cite{minaee2021image}. The emergence of encoder-decoder-based architectures, such as Fully Convolutional Networks (FCNs) \cite{long2015fully} and U-shaped structures, initially proposed as U-Net \cite{ronneberger2015u} and its variations, has contributed significantly to the notable success of CNNs in the medical image segmentation field. In these architectures, skip connections were introduced to improve fine-grained segmentation performance to aggregate multi-stage features to the decoder branch for generating high-resolution segmentation maps. To further improve the success of the U-Net model, the variant structures such as UNet++ \cite{zhou2018unet++}, Attention U-Net \cite{oktay2018attention}, Dense U-Net \cite{wang2019dense}, V-Net \cite{milletari2016v}, and nnU-Net \cite{isensee2021nnu}, an automatically optimized U-Net architecture, have been proposed. Despite demonstrating remarkable performance, CNN-based methods require deeper architectures, thus high trainable parameters to learn long-range dependencies among pixels due to the limited receptive field and the inductive bias of the convolution operation \cite{dosovitskiy2020image, liu2021swin}.\\
To fill the semantic gap between the local and global features, the attention mechanism was introduced to the CNN-based architectures to enhance the long-range understanding of the CNN modules by concentrating on relevant spatial regions and feature channels. Various types of attention mechanisms have been introduced and implemented in the models \cite{lin2022ds, valanarasu2021medical, sinha2020multi, gao2021utnet, wang2018deep, oktay2018attention}. In their research, Wang et al. \cite{wang2018deep} employed attention modules at various resolutions to incorporate deep attention features (DAF) with a global context, enhancing the precision of prostate segmentation in ultrasound images. Oktay et al. \cite{oktay2018attention} introduced the Attention U-Net, which incorporates attention gates into the U-Net architecture to capture global contexts by combining the skip and decoder branch connections. Sinha et al. \cite{sinha2020multi} proposed a multi-scale approach to merge semantic information from different levels and include self-attention layers, named position and channel attention modules, to aggregate relevant contextual features systematically. While attention modules enhance important features, there is still a limitation in effectively bridging semantic gaps between local and global features for CNNs.

The Transformer \cite{vaswani2017attention} model, originally a prevalent choice for natural language processing (NLP) tasks, has recently gained extensive usage in vision-based applications \cite{dosovitskiy2020image} to address the weaknesses of the CNN models. The fundamental concept is to employ the multi-head self-attention (MSA) mechanism to effectively capture long-range dependencies between the sequence of tokens generated from the non-overlapping image patches. The first attempt of the vision-based transformers, named Vision Transformer (ViT) \cite{dosovitskiy2020image}, demonstrates performance on par with CNN-based methods; however, it demands substantial amounts of data for effective generalization and is burdened by quadratic complexity. In more recent developments, hierarchical vision transformers, named Swin Transformer \cite{liu2021swin}, incorporating window-based attention, and the Pyramid Vision Transformer (PVT) \cite{wang2021pyramid}, featuring spatial reduction attention have emerged to mitigate computational costs. Swin-Unet \cite{cao2022swin} proposes a Swin Transformer-based encoder and decoder architecture to capture global features but lacks local contextual information to generate fine-grained edge segmentation. Studies that integrate both local and global dependencies through hybrid methods have also been proposed. For instance, TransUNet \cite{chen2021transunet} presents a ViT-based encoder branch and CNN-based decoder branch as a hybrid approach to extract local and global features. Similarly, Swin UNETR \cite{hatamizadeh2021swin} and HiFormer \cite{heidari2023hiformer} introduce the Swin Transformer in their architecture to reduce the complexity and improve the global feature extraction to perform 3D segmentation. 

Hybrid short and long-range relation extraction capabilities of the CNN-transformer approaches have made them more effective for medical image segmentation compared to standalone CNN-based studies \cite{tragakis2023fully, yuan2023effective}. Nevertheless, there are still various limitations that need to be addressed. Firstly, the multi-scale features from the hierarchical encoder branch are rarely utilized, which results in the loss of fine-grained border information in the segmentation map. Secondly, the previous hybrid architecture designs did not properly balance complexity and accuracy as transformer blocks were introduced in high-level stages, increasing trainable parameters. Finally, the attentive features are rarely fused with the decoder branch, reducing the precise mask generation. Therefore, our motivation behind GLIMS is to address the challenges of computational scalability and the efficient utilization of hybrid architectures for feature consistency. We achieve this by utilizing depth-wise multi-scale feature extractors and efficient transformer layers, incorporating additional attention mechanisms to improve the localization of mask generation, an aspect that has been overlooked in previous literature.

In our work, we propose a novel attention-guided lightweight multi-scale hybrid CNN-transformer-based segmentation model, GLIMS, by utilizing the locality attribute of CNNs, and the long-range relation extraction ability of transformers to process medical scans effectively. The encoder and decoder branches are designed to have CNN layers at high levels to increase the local feature extraction from high-resolution slices and transformer layers at lower levels to extract long-range dependencies by reducing the complexity and improving the feature consistency. The encoder and decoder branches are symmetrically composed of the proposed \textbf{D}ilated Feature \textbf{A}ggregator \textbf{C}onvolutional \textbf{B}locks (DACB) to benefit from the locality of CNNs by increasing their global context representation abilities via dilated kernels. A Swin Transformer-based bottleneck is incorporated to improve global feature extraction by hybridly fusing the local-global features extracted from the DACB blocks. To further guide the model in localizing the interested regions during mask generation, parallelly structured \textbf{C}hannel and \textbf{S}patial-Wise \textbf{A}ttention \textbf{B}locks (CSAB) are included in the skip connections, which transfers the high-resolution features. Our proposed model, GLIMS, improves local-global feature connections in a lightweight form and outperforms its counterparts by reducing complexity and improving the segmentation accuracy based on Dice Score and 95\% Hausdorff Distance on BraTS2021 \cite{bakas2017advancing} and BTCV \cite{gibson_2018_1169361} medical image segmentation datasets. The novel design of GLIMS placed it in the top-5 ranking in the validation phase of BraTS2023 Adult Glioblastoma Segmentation Challenge \cite{menze2014multimodal}. The main contributions of the study can be summarized as:

\begin{itemize}
    \item We introduce GLIMS, an attention-guided lightweight multi-scale hybrid network for the volumetric medical image segmentation task, surpassing previous CNN and hybrid models on 3D glioblastoma and multi-organ segmentation datasets.
    
    \item GLIMS achieves notable improvements with a Dice Score of 92.14\% and 84.50\% on BraTS2021 and BTCV, respectively, outperforming state-of-the-art models by maintaining efficiency with 47.16M trainable parameters and 72.30G FLOPs.

    \item The proposed Dilated Feature Aggregator Convolutional Blocks module utilizes the locality of CNNs and long-range extraction performance of dilated kernels via depth-wise operations to establish effective feature extraction with reduced complexity.

    \item The proposed Channel and Spatial-Wise Attention Blocks module localizes the region of interest in the 3D scans to guide the decoder branch in a learnable manner, improving segmentation accuracy.

    \item The Swin Transformer-based bottleneck strengthens the connection between locally extracted features and long-range dependencies, enhancing the segmentation of large regions and achieving fine-grained segmentation performance on edges.
\end{itemize}

The remainder of the paper is organized as follows: First, a detailed background on models for volumetric semantic segmentation will be provided. Next, the design motivations behind GLIMS and its proposed modules will be explained in detail. Afterward, the experimental results on BraTS2021 and BTCV datasets will be presented and analyzed both quantitatively and qualitatively. In addition, a comprehensive set of ablation studies will also be provided and discussed. Finally, the last section will conclude the paper.

\section{Related Works}

This section will present an overview of the literature concerning medical image segmentation tasks that integrate CNNs, explore widely adopted vision transformer models in segmentation architectures, and highlight hybrid approaches combining CNNs and vision transformers for volumetric medical image segmentation tasks.

\subsection{CNN-based Segmentation Models}

CNNs have been widely used for medical image segmentation \cite{minaee2021image}. The U-Net architecture, introduced in \cite{ronneberger2015u}, is one of the most popular deep segmentation models. Due to its simplicity and superior performance, various U-Net-based methods continue to emerge, employing diverse strategies to enhance feature extraction by including the integration of different residual and dense blocks. Diakogiannis et al. \cite{diakogiannis2020resunet} propose Res-UNet, which incorporates residual connections, atrous convolutions, and pyramid scene parsing pooling in the encoder and decoder components. Cai et al. \cite{cai2020dense} utilizes dense concatenation to increase the depth of the network architecture and facilitate feature reuse in Dense-UNet, including four expansion modules, each comprising four down-sampling layers for feature extraction. Stoyanov et al. \cite{stoyanov2018deep} propose U-Net++ that forms nested hierarchical dense skip connection pathways between the encoder and decoder to mitigate the semantic gap present in the learned features. As an improved version of U-Net++, Huang et al. \cite{huang2020unet} utilize full-scale skip connections and deep supervision in UNet3+, combining low-level details with high-level semantics from feature maps at different scales and employing deep supervision for learning hierarchical representations from fully aggregated feature maps. Tragakis et al. \cite{tragakis2023fully} propose a fully convolutional network named FCT, which incorporates multi-head self-attention modules at each stage of the model by utilizing the convolutional embeddings. Gu et al. \cite{gu2020net} integrate spatial, channel, and scale attention mechanisms in CA-Net to focus on the most relevant regions, features, and object sizes, respectively, for image segmentation by enhancing both accuracy and explainability.

This architectural paradigm has also found application in 3D medical image segmentation by 3D-Unet \cite{cciccek20163d} and V-Net \cite{milletari2016v}, which convert the 2D convolution layers with 3D kernels to perform voxel-wise semantic segmentation. As more advanced approaches, Roy et al.  \cite{roy2023mednext} propose MedNeXt, a large kernel segmentation network for medical image segmentation, utilizing a ConvNeXt \cite{liu2022convnet} 3D Encoder-Decoder Network, Residual ConvNeXt up and downsampling blocks, and a technique to iteratively increase kernel sizes for enhanced contextual and global feature extraction. Lastly, Cao et al. \cite{cao2023mbanet} introduce MBANet, which uses a multi-branch 3D shuffle attention module in the encoder layer to group feature maps by channel, divide them into shuffled features, and apply channel and spatial-wise attention modules to localize attention to region of interest.

\subsection{Vision Transformers}

The Transformer architecture, originally proposed for Natural Language Processing (NLP) tasks \cite{vaswani2017attention}, has demonstrated state-of-the-art results in the computer vision domain since the introduction of Vision Transformers \cite{dosovitskiy2020image}. ViT excels in image classification performance by leveraging self-attention mechanisms to facilitate the learning of global information. However, ViT's limitation lies in its reliance on pre-training on a large dataset. To overcome this, Touvron et al. \cite{touvron2021training} proposed data-efficient image transformers, DeiT, with innovative multiple training strategies that effectively utilize the vision transformer models. ViT exhibits promising results in image classification; however, its architecture, which involves dividing the input image into equal-sized, non-overlapping patches to generate a sequence of tokens, leads to a quadratic increase in complexity based on the input image size. This attribute makes it impractical as a general-purpose backbone, particularly for medical tasks where input images typically exhibit high resolution and multiple spatial dimensions. To overcome this, Liu et al. \cite{liu2021swin} proposed Swin Transformer, an efficient hierarchical Vision Transformer. It utilizes shifted windows multi-head self-attention (SW-MSA) for a local self-attention computation to improve the local context understanding of the model and reduce the quadratic complexity to linear. Other studies have also proposed attention mechanisms to consider local and global interactions to reduce the semantic gap \cite{wang2021crossformer, lin2022cat, wu2021cvt}. Nevertheless, owing to the computational demands of transformer models and the better representation ability of the CNNs for the local features, hybrid approaches have begun to be introduced \cite{khan2022transformers}.

\begin{figure*}[t!]
    \centering
    \includegraphics[width=\linewidth]{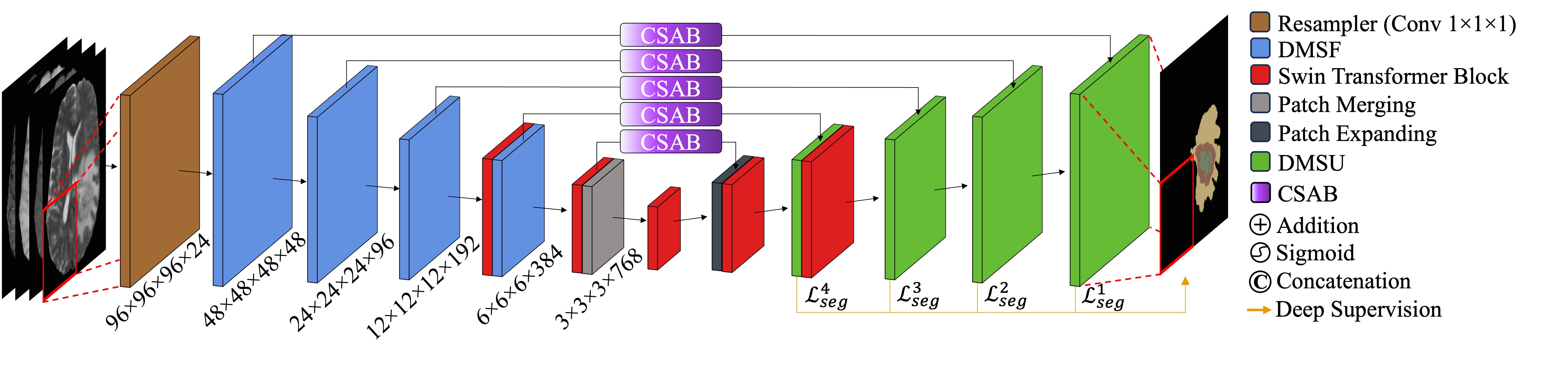}
    \caption{The proposed architecture of GLIMS. The model utilizes a patch-based segmentation approach, where each volume is resampled with the initial layer to expand the feature dimension to \textit{24}. The resampled features are processed with a sequence of depth-wise multi-scale convolutional blocks, DMSF, and fused with the transformer bottleneck. Features from each stage of the encoder branch are then processed with an Attention-Guidance module, CSAB, to localize the features. Finally, the attention maps are concatenated with the up-sampled features by DMSU, and the segmentation mask is predicted. In addition, deep supervision is performed to effectively supervise the model's convergence.}
    \label{fig:Model}
\end{figure*}

\begin{figure}
    \centering
    \includegraphics[width=\linewidth]{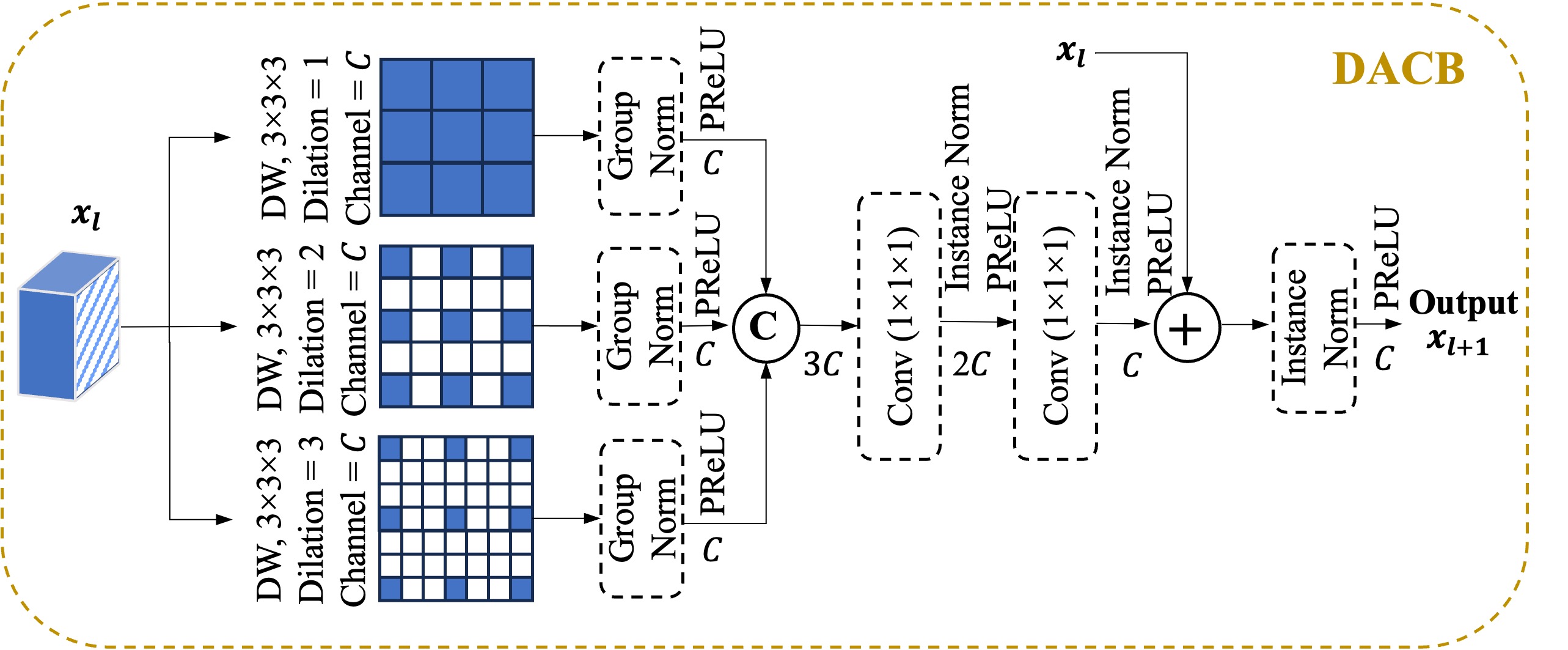}
    \caption{The proposed Dilated Feature Aggregator Convolutional Block, DACB, prevents the loss of border information at low resolution, enabling the extraction of both global and local features at higher levels using dilated kernels.}
    \label{fig:Model_2}
\end{figure}

\subsection{Hybrid Segmentation Models}

By integrating the local representation capabilities of CNNs with the long-range dependency modeling of transformers, hybrid models have demonstrated promising results in enhancing computer vision task performance while concurrently reducing model complexity. As one of the pioneering hybrid medical image segmentation models, TransUNet \cite{chen2021transunet} enhances global context by treating image features as sequences through the use of ViT, while also effectively leveraging low-level CNN features. UNETR \cite{hatamizadeh2022unetr} processes 3D volumetric patches from the input scan, embedding them through transformer-based encoders, and employs a CNN-based decoder for performing 3D segmentation. To mitigate the complexity of the model when handling 3D volumes, Swin UNETR \cite{hatamizadeh2021swin} employs a Swin Transformer on the encoder branch, while the segmentation masks are generated using a CNN-based decoder. TransBTS \cite{wang2021transbts} combines 3D-CNN with transformer layers, placing transformers in the bottleneck of the architecture. nnFormer \cite{zhou2021nnformer} incorporates Local Volume-based Multi-head Self-attention (LV-MSA) and Global Volume-based Multi-head Self-attention (GV-MSA) on top of convolutional layers to construct feature pyramids for learning representations on both local and global 3D volumes. Heidari et al. \cite{heidari2023hiformer} introduce HiFormer, a novel hybrid CNN-transformer-based method that combines the global features obtained from a Swin Transformer module with the local representations of a CNN-based encoder. Then, using a double-level fusion layer module, they combine multi-scale representation for finer segmentation performance. He et al. \cite{he2023h2former} propose H2Former, a hierarchical hybrid semantic segmentation model that blends the local information of convolutional neural networks (CNNs), multi-scale channel attention features, and long-range features of Transformer within a unified block. Liu et al. \cite{liu2022phtrans} propose a hybrid architecture named PHTrans, which parallelly combines Transformer and CNN in main building blocks to create hierarchical representations from global and local features. It then adaptively aggregates these representations, with the aim of fully exploiting their strengths to obtain better segmentation performance. 

On the other hand, the Large Vision Model (LVM) is a developing field that has found applications in various tasks in computer vision. These models can be trained on large datasets that comprise both textual and visual data on a main task and adapted into downstream tasks, such as segmentation \cite{liu2024sora}. In terms of the segmentation task, various methods have previously utilized visual prompts such as point, bounding box, or mask to predict the target segmentation \cite{luddecke2022image,wang2023seggpt,kirillov2023segment,zou2024segment,hong2023spectralgpt}. Yet, LVMs suffer from specific issues for medical tasks: non-domain-specific knowledge and high complexity \cite{awais2023foundational}. Medical imaging often lacks enough data for training image segmentation models, especially when dealing with volumetric data. As a result, generic datasets are used to train these models. However, due to the large size of these datasets, the models are usually developed with high capacity by incorporating CNN and ViT models in a hybrid approach and may be difficult to train on a standard system.

Additionally, according to the literature, existing models suffer from limited incorporation of local features, reducing the clarity of inter-class regions in output masks. Moreover, their complexity makes them data-dependent, posing challenges in low-data scenarios, typical in the medical field. This motivated us to develop a lightweight, multi-scale feature extraction and attention-guided hybrid method for efficient training and robust inter- and intra-class region segmentation.

\section{Methods}

This section gives the overall structure of the proposed GLIMS model for 3D medical image segmentation, as illustrated in Figure \ref{fig:Model}. The proposed model incorporates Swin Transformer architecture to effectively capture the global features in 3D medical scans and utilize \textbf{D}epth-Wise
\textbf{M}ulti-\textbf{S}cale \textbf{F}eature Extraction (DMSF) modules to further improve the modeling of the long-range dependencies by integrating the local representations. Furthermore, the mask generation in the decoder branch is supported with attention guidance via \textbf{C}hannel and \textbf{S}patial-Wise \textbf{A}ttention \textbf{B}locks (CSAB) to improve inter and intra-class uniformity in segmentation.

\begin{figure}
    \centering
    \includegraphics[width=\linewidth]{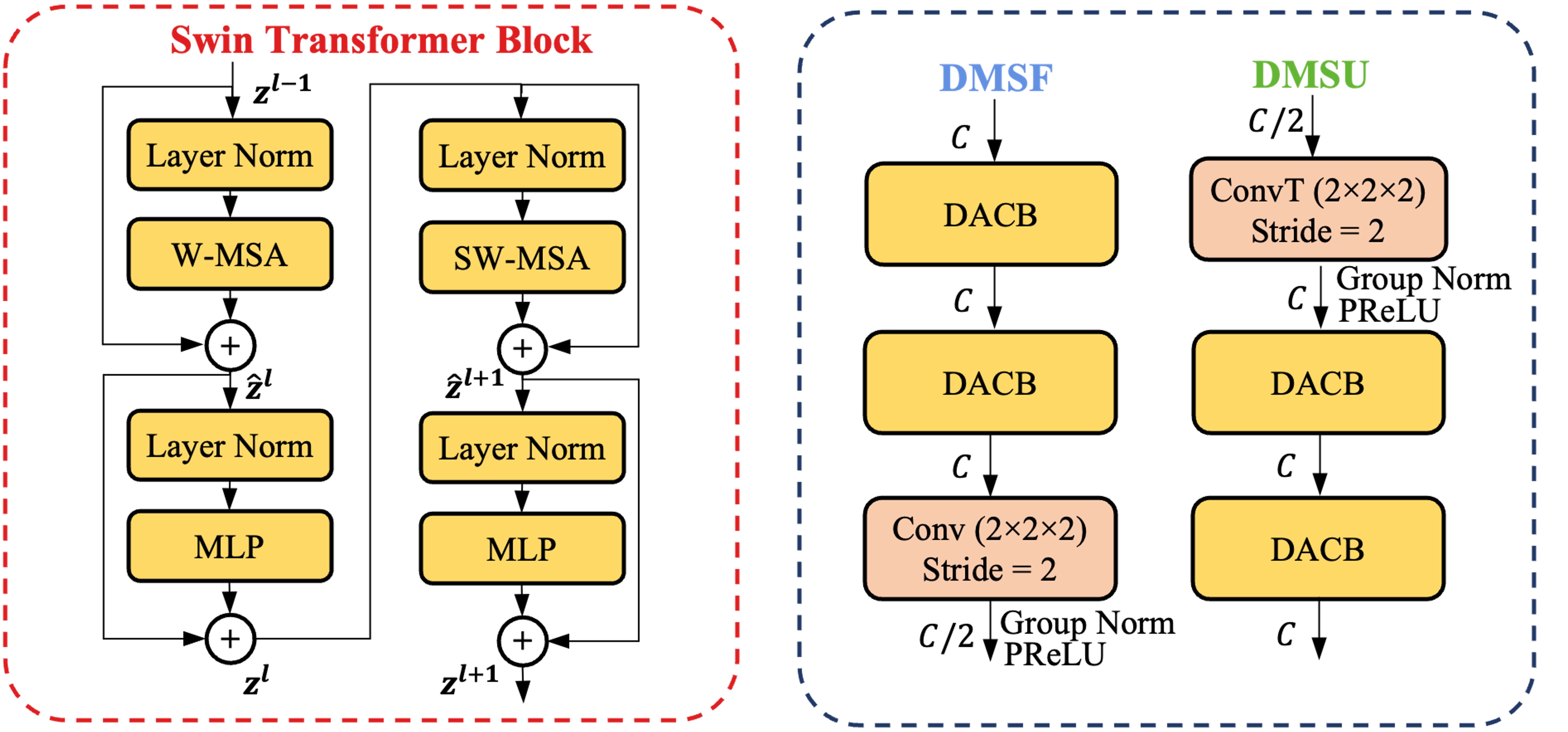}
    \caption{The Swin Transformer Block architecture on the left and the proposed DMSF and DMSU blocks on the right. The DMSF and DMSU modules comprise two consecutive DACB blocks and each module has an individual downsampling or upsampling convolutional block.}
    \label{fig:Model_4}
\end{figure}

\subsection{Dilated Feature Aggregating Encoder}

The encoder component of GLIMS integrates nested CNN modules and Swin Transformer layers in a hybrid structure to enhance retrieved features globally and locally. Within each layer, CNNs are employed in a depth-wise configuration to mitigate the increase of trainable parameters arising from 3D kernels. Additionally, feature extraction is performed in a dilated structure, thereby harnessing the capability to capture fine-grained multi-region edge information to be contained in the generated mask. This approach is designed to integrate local and global information and effectively represent the structure of the regions.

The proposed model adopts a patch-based methodology specifically designed for small GPUs to optimize efficiency on limited graphical processing units. In this approach, a random patch is sampled from the entire input scan and given to the model during each iteration. This reduces the memory requirements during training and serves as an additional augmentation process. Consequently, the model's input size is defined as $X \in \mathbb{R}^{H \times W \times D \times S}$, where $H$, $W$, and $D$ are all set to 96. The initial input undergoes resampling through a $1 \times 1 \times 1$ point-wise convolution, resulting in a depth of $S$, which is 24. Within each layer of the encoder branch, a reduction is performed in the spatial resolution of the feature matrix by a factor of two. Simultaneously, there is a twofold increase in the channel resolution. Every processed feature map is connected with the decoder branch, facilitating the transfer of both local and global features. In each layer, the feature maps are processed with the DMSF blocks, which include two consecutive \textbf{D}ilated Feature \textbf{A}ggregator \textbf{C}onvolutional \textbf{B}locks (DACB) module, shown in Figure \ref{fig:Model_2}. In order to leverage both local and global feature extraction, three parallel layers of $3 \times 3 \times 3$ dilated CNNs are employed. These layers generate new feature representations denoted as $Z_d$, where $d \in \{1,2,3\}$ representing each dilation factor. For the aggregation of features within each $Z_d$, the respective feature maps are concatenated and subjected to further processing through two $1 \times 1 \times 1$ CNN layers. This processing gradually reduces the channel size, resulting in the output denoted as $Z_f$. Subsequently, the input to the module, $x_l$, is summed with $Z_f$ to preserve gradient flow towards higher layers. The instance-normalized output is then forwarded to the subsequent modules. This design allows for the retrieval of global context within the patch without significant spatial dimension reduction, enabling the preservation of edge details crucial for fine-grained segmentation output. 

\begin{figure} [t!]
    \centering
    \includegraphics[width=\linewidth]{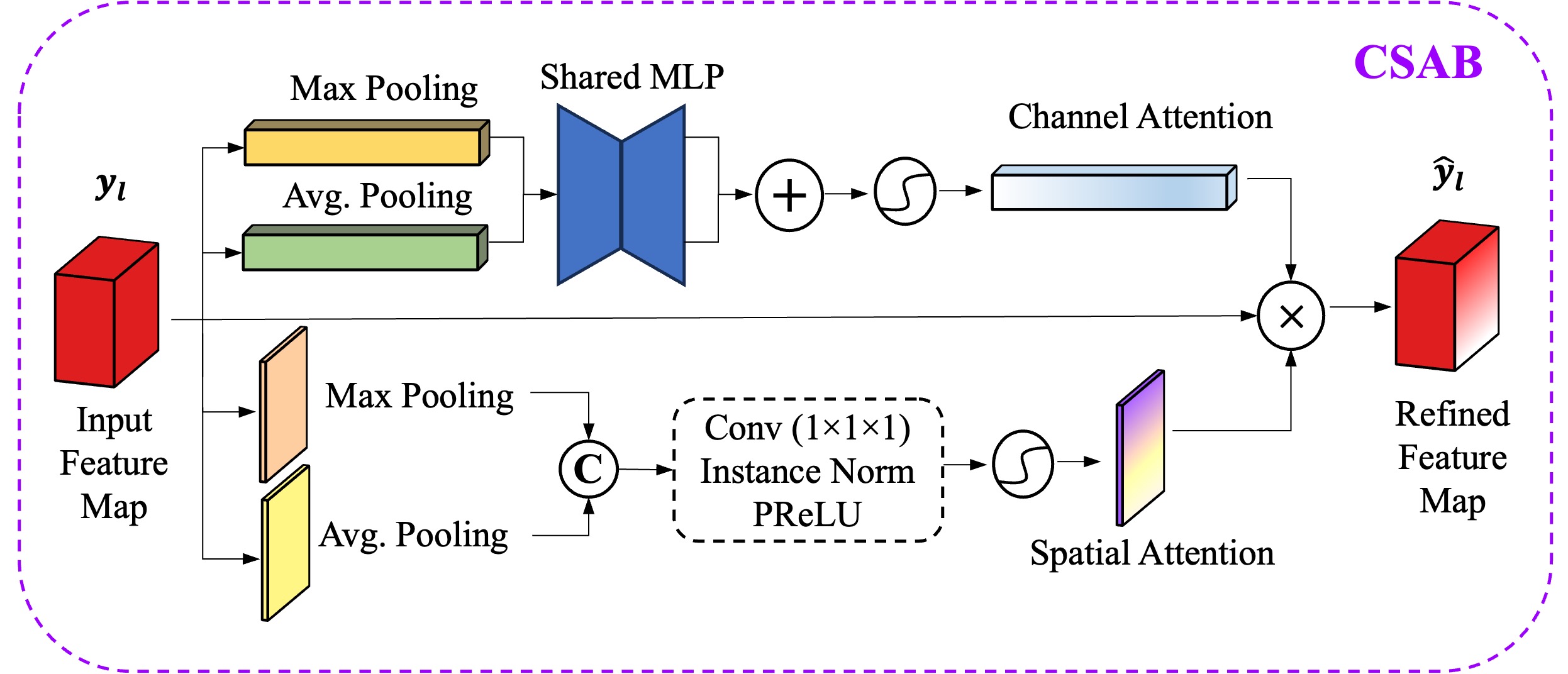}
    \caption{The proposed attention module, CSAB, fuses the local and global attention of the input feature map in parallel branches and generates a refined feature map with localized attention on the region of interest.}
    \label{fig:Model_3}
\end{figure}

\subsection{Swin Transformer Bottleneck}

The lower levels of the proposed model are designed as a hybrid approach of CNNs and transformer blocks, aiming to maintain concurrent fine-grained local and global feature extraction. The primary motivation behind this hybrid design is that while dilated kernels contribute to global context extraction, they exhibit limitations in constructing long-range dependencies when compared to transformers. On the other hand, although they excel in global feature extraction, transformers come at the expense of higher trainable parameters with large feature maps in spatial dimensions compared to CNNs.

The fundamental structure of ViT, \cite{dosovitskiy2020image}, comprises multi-head self-attention and multi-layer perceptron (MLP) layers to process the sequence-of-tokens with the self-attention mechanism. A significant challenge associated with ViT lies in its quadratic complexity, rendering it inefficient for high-resolution computer vision tasks such as image segmentation, especially for 3D medical scans. To address this limitation, the Swin Transformer \cite{liu2021swin}, as shown in Figure \ref{fig:Model_4}, introduced W-MSA to perform self-attention within each window individually and SW-MSA to have cross-window connections between the local windows to improve long-range links, which are used in the deeper encoder, decoder, and bottleneck parts of GLIMS.

The input features to the transformer blocks are first partitioned with a patch size of $2 \times 2 \times 2$, resulting in tokens of dimensions $\left[\frac{H}{2}\right] \times \left[\frac{W}{2}\right] \times \left[\frac{D}{2}\right]$ within a local window of size $7 \times 7 \times 7$. These generated patches are summed with learnable positional embeddings, configured in the shape of $\left[\frac{H}{2}\right] \times \left[\frac{W}{2}\right] \times \left[\frac{D}{2}\right] \times C$, where $C$ represents the hidden size of the current layer. Self-attention modules are implemented on non-overlapping embedding windows to reduce the parameter size. To execute attention at the transformer level $l$, the 3D tokens are evenly partitioned into $\left[\frac{H'}{M}\right] \times \left[\frac{W'}{M}\right] \times \left[\frac{D'}{M}\right]$, where $M \times M \times M$ denotes the window resolution, and $H'$, $W'$, and $D'$ represent the current shape of the feature matrix in height, width, and depth, respectively. In the subsequent layer $l+1$, the patches undergo a shift to capture local context. This patch shifting enables each patch to attend to its adjacent counterparts, allowing it to gather information from the surrounding features. The shift operation ensures overlapping receptive fields of the patches, facilitating the model in effectively capturing local and global information. To accomplish this, the windows are shifted by $\left(\left[\frac{M}{2}\right], \left[\frac{M}{2}\right], \left[\frac{M}{2}\right]\right)$ voxels. Subsequently, patch-merging is employed, involving the concatenation of 2 × 2 groups of adjacent patches to increase embedding dimensions from $C$ to $2C$ and reducing the spatial resolution by 2. The patch-expanding module is executed in the decoder branch to achieve the reverse operation.

\begin{equation}
\begin{aligned}
\hat{z}_l &= \text{W-MSA}(\text{LN}(z_{l-1})) + z_{l-1} \\
z_l &= \text{MLP}(\text{LN}(\hat{z}_l)) + \hat{z}_l \\
\hat{z}_{l+1} &= \text{SW-MSA}(\text{LN}(z_l)) + z_l \\
z_{l+1} &= \text{MLP}(\text{LN}(\hat{z}_{l+1})) + \hat{z}_{l+1}
\end{aligned}
\label{eq:Swin}
\end{equation}

The process is described by Equation \ref{eq:Swin}, with W-MSA and SW-MSA as windowed and shifted window multi-head self-attention modules, respectively, while LN and MLP represent Layer Normalization and Multi-Layer Perceptron. 

\begin{figure} [t]
    \centering
    \includegraphics[width=\linewidth]{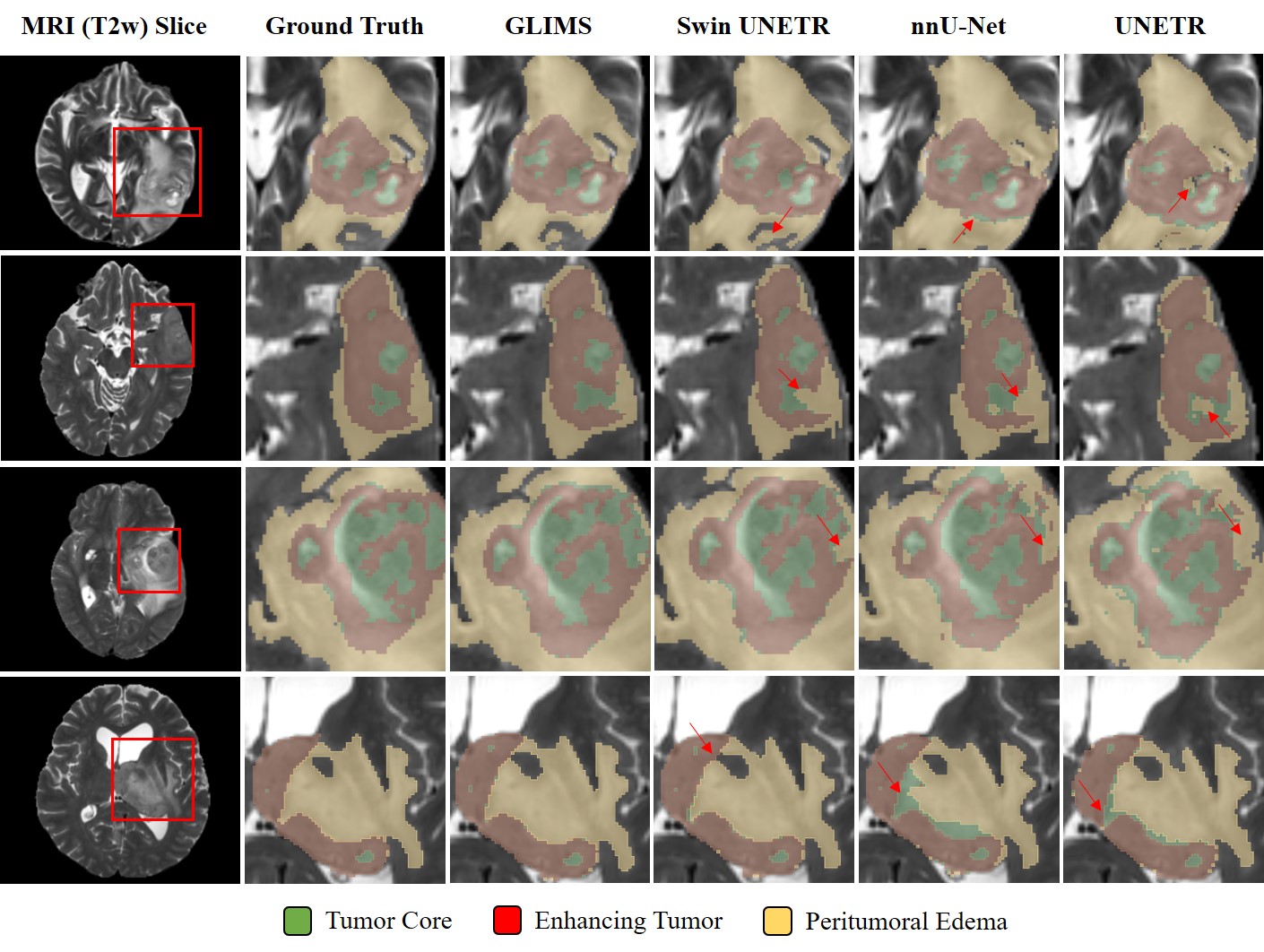}
    \caption{Qualitative results on the BraTS2021 dataset. Each row represents a different patient, and from left to right, the Ground truth mask and the prediction results of GLIMS are provided, respectively. The \textcolor{green}{green}, \textcolor{red}{red} and \textcolor{yellow}{yellow} colored regions represent Tumor Core (TC), Enhancing Tumor (ET) and Peritumoral Edema (ED) regions, respectively.}
    \label{fig:BraTS_Seg}
\end{figure}

\subsection{Channel and Spatial-Wise Attention Block}

As the depth of the neural network structure increases, the model gains the capacity to learn intricate transformations and accommodate more complex feature inputs. However, guiding the model to emphasize the intuitive features and to suppress unnecessary voxel-wise signals may enhance segmentation performance. This is achieved by forwarding the retrieved attention to the decoder branch to guide the mask operation via fused attention. To improve the representation of the features, we introduce parallel-branched \textbf{C}hannel and \textbf{S}patial-Wise \textbf{A}ttention \textbf{B}locks (CSAB), as given in Figure \ref{fig:Model_3}. The CSAB module refines input feature maps by independently enhancing or inhibiting features in the channel and spatial dimensions, thus enhancing global features and highlighting the salient features of local regions to achieve more effective feature guidance during segmentation. 

The input feature map, $y_l$, is processed through max pooling and average pooling operations, aggregating features separately in the channel and spatial dimensions. Afterward, the spatially reduced features are processed via a weight-shared MLP model to introduce learnable attention into the new feature embeddings. The resulting embeddings are then summed to highlight high-attention locations and suppress low-attention regions. Lastly, the generated feature embedding is normalized to the range of zero and one via sigmoid function. Simultaneously, pooling operations are repeated in the channel-wise processing branch, concatenating the resulting outputs. This concatenated output is then channel-wise reduced through a $1 \times 1 \times 1$ CNN layer and normalized using sigmoid layer. To refine the initial feature map, $y_l$, the generated attention maps are multiplied, yielding the refined feature map $\hat{y}_l$.

\begin{figure} [t]
    \centering
    \includegraphics[width=\linewidth]{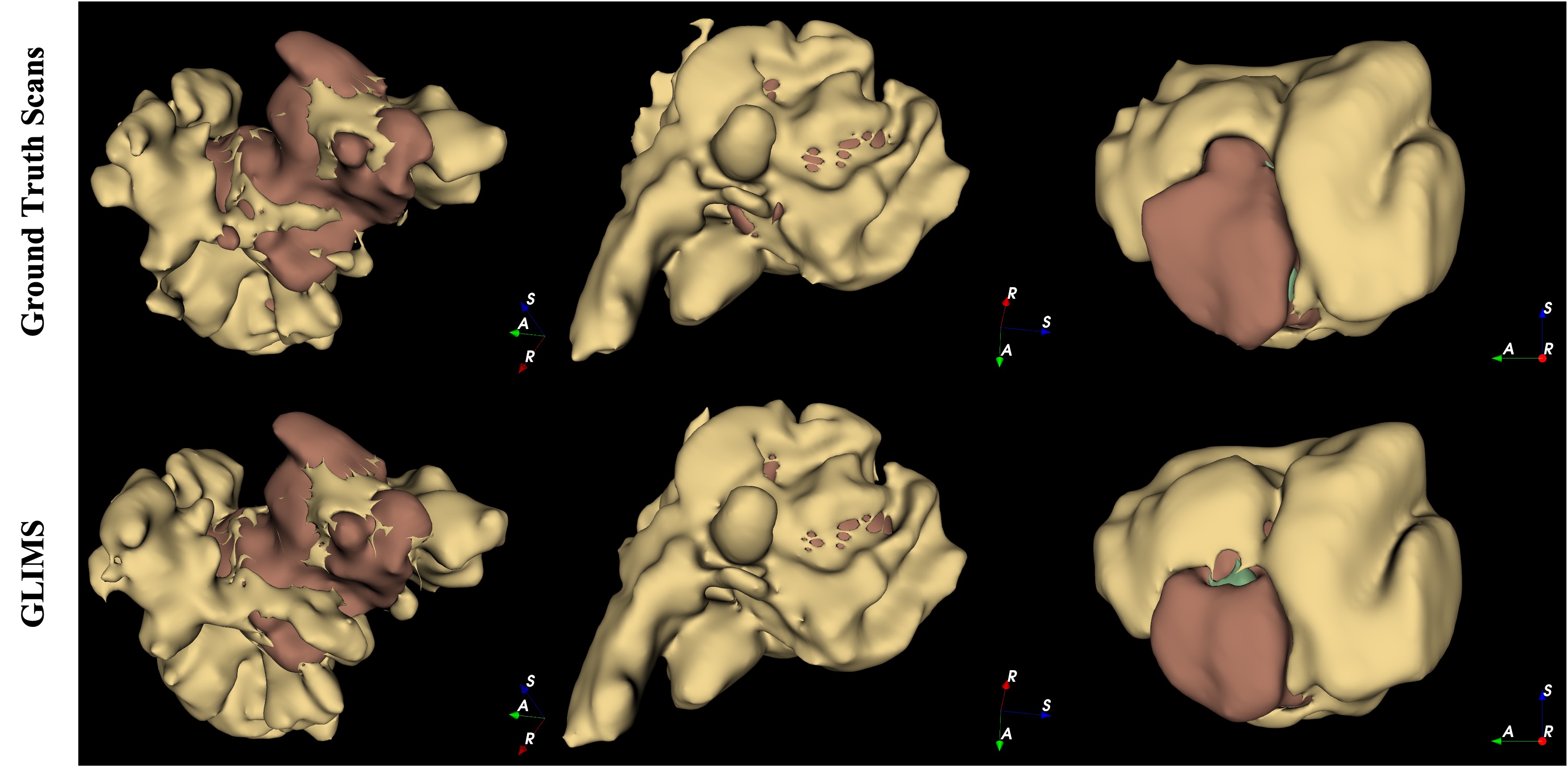}
    \caption{Qualitative 3D segmentation results on the BraTS2021 dataset. Each column represents a new patient. The first row shows the 3D mesh of the ground truth image, while the second row shows the predictions of GLIMS.}
    \label{fig:BraTS_3D}
\end{figure}

\subsection{Attention Guiding Multi-Scale Decoder}

The locally and globally enhanced features from the bottleneck are up-sampled via the \textbf{D}epth-Wise \textbf{M}ulti-\textbf{S}cale \textbf{U}psampling (DMSU) module, as presented in Figure \ref{fig:Model_4}, by a factor of two in each level. In each module, the feature map is processed by a $2 \times 2 \times 2$ 3D transpose CNN module to decrease the channel size from $2C$ to $C$, and two consecutive DACB modules are followed. In each level, the input features are up-sampled by a factor of two and concatenated with the $\hat{y}_l$ features given from the CSAB module from each hierarchical layer of the encoder branch. The new channel dimension as $2C$ is again reduced into $C$ via a $1 \times 1 \times 1$ CNN layer to be fused with the following higher level of the refined features. The fusion of refined features to generate a segmentation mask helps the model concentrate on predicting the region of interest, avoiding unnecessary segmentation in irrelevant areas. Thus, the model optimizes itself in a learnable manner to integrate the attention during the mask generation phase.


\subsubsection{Deep Supervision}

To leverage predictions from lower layers, we applied deep supervision in GLIMS inspired by the nnU-Net \cite{isensee2021nnu} approach. Deep supervision is a technique that involves computing the loss function not only from the last layer but also incorporating logits from intermediate layers. This approach includes training CNNs with multiple intermediate supervision signals, leading to improved segmentation results. In traditional training, the network is trained end-to-end with a single final output, making identifying and correcting errors at different stages challenging. However, by introducing intermediate supervision, additional loss values are retrieved at multiple network layers, allowing the model to learn more discriminative and informative features. 

\begin{equation}
    \mathcal{L}_{seg} = 1 - \frac{2}{K}\sum_{k=1}^{K}\frac{\sum_{i=1}^{N} y_{i,k} p_{i,k}}{\sum_{i=1}^{N} y_{i,k}^2 + \sum_{i=1}^{N} p_{i,k}^2}
    -\sum_{k=1}^{K}\sum_{i=1}^{N} y_{i,k} \log(p_{i,k})
\label{eq:DiceCELoss}
\end{equation}

\begin{equation}
    \mathcal{L}_{DS} = \mathcal{L}^1_{seg} + \frac{1}{2}\mathcal{L}^2_{seg} \\
    + \frac{1}{4}\mathcal{L}^3_{seg} + \frac{1}{8}\mathcal{L}^4_{seg}
\label{eq:DS}
\end{equation}

We utilized the equal-weighted combination of Dice Loss and Cross-Entropy Loss, $\mathcal{L}_{seg}$, as shown in Equation \ref{eq:DiceCELoss}, to optimize GLIMS. In the equation, $K$ represents the total number of classes, $N$ is the voxel count, $y$ denotes the ground truth labels, and $p$ is the predicted one-hot classes. The employed deep-supervision loss function, $\mathcal{L}_{DS}$, to be used for the back-propagation is presented in Equation \ref{eq:DS}, where each $L^{i}_{seg}, i \in \{1,2,3,4\}$ represents the loss values corresponding to the combination of Dice Loss and Cross-Entropy for level $i$. This combination is also depicted in Figure \ref{fig:Model}. In the calculation of $\mathcal{L}_{DS}$, the weights assigned to $\mathcal{L}_{seg}$ decrease for deeper layers, with shallower layers having higher weights.

\begin{figure} [t]
    \centering
    \includegraphics[width=\linewidth]{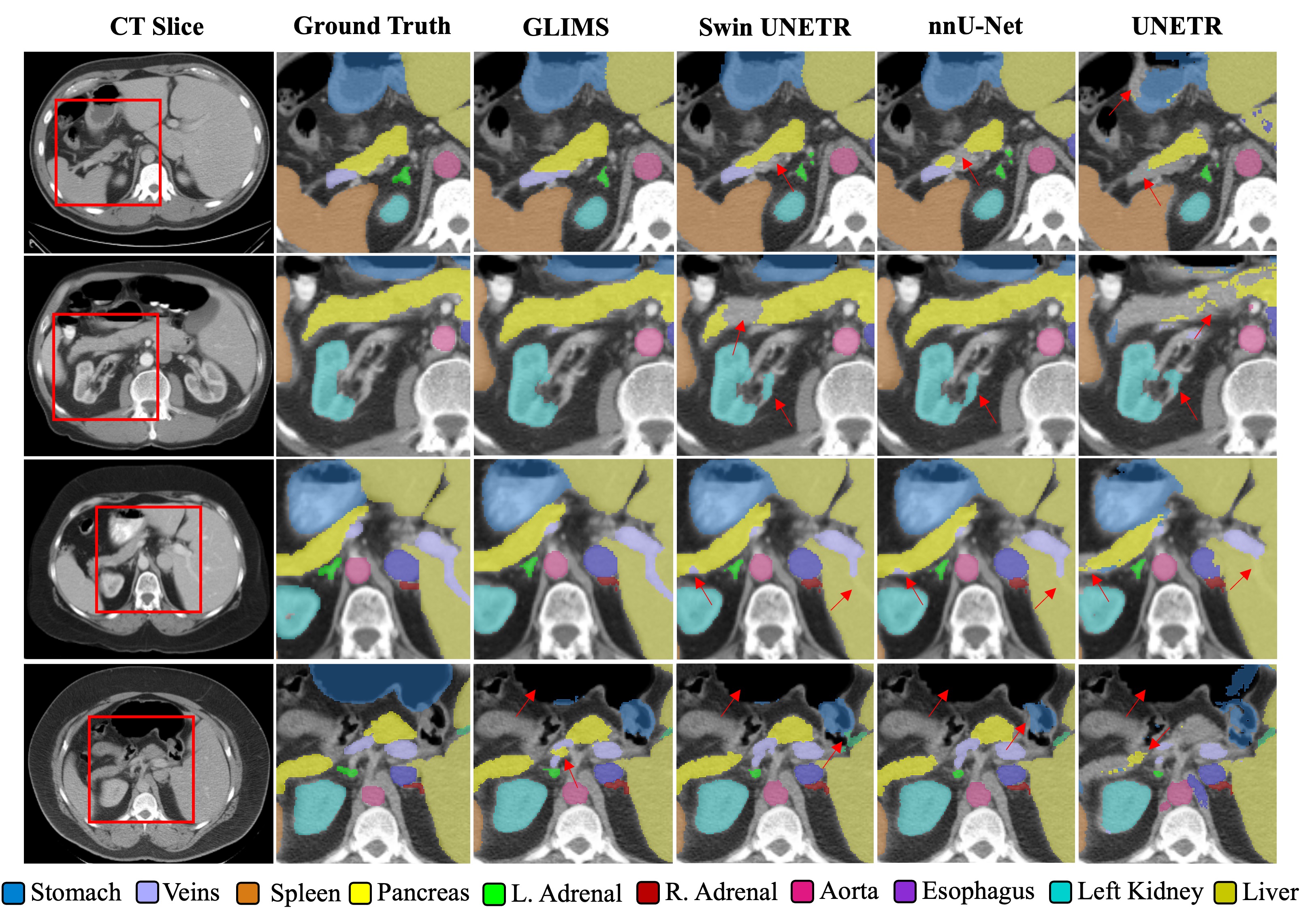}
    \caption{Qualitative results on the BTCV dataset. Each row represents a different patient, and from left to right, the Ground truth mask and the prediction results of GLIMS, Swin UNETR, nnU-Net, and UNETR are provided, respectively.}
    \label{fig:BTCV_Seg}
\end{figure}

\section{Experiments \& Discussion}

This section presents quantitative and qualitative results for two volumetric segmentation tasks: Brain MRI tumor segmentation and multi-organ CT segmentation. Furthermore, the results of ablation studies will be provided and a detailed discussion of the findings will be presented.

\begin{figure} [t]
    \centering
    \includegraphics[width=\linewidth]{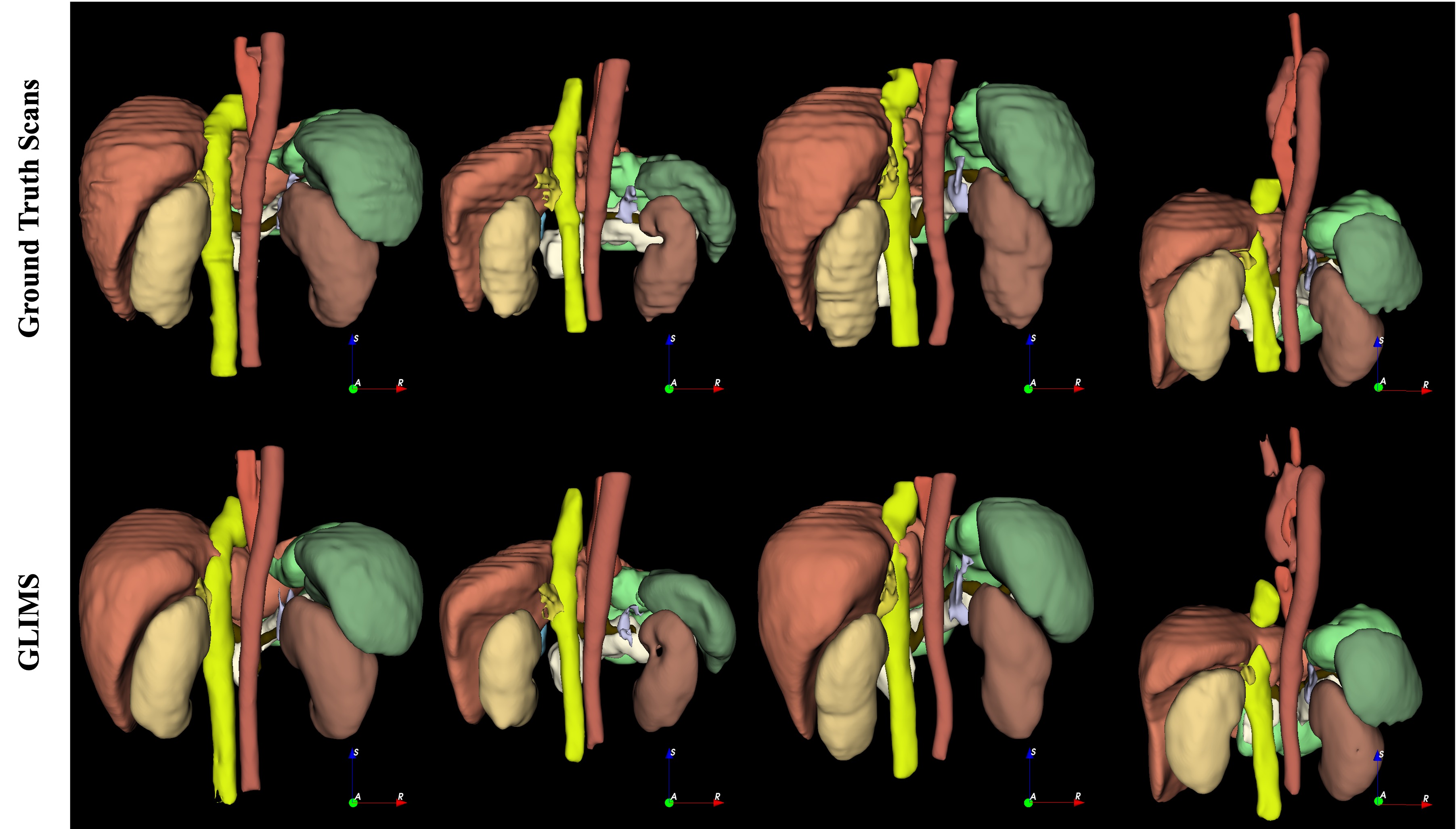}
    \caption{Qualitative 3D segmentation results on the BTCV dataset. Each column represents a new patient. The first row shows the 3D mesh of the ground truth image, while the second row shows the predictions of GLIMS.}
    \label{fig:BTCV_3D}
\end{figure}

\subsection{Datasets}
To evaluate the performance of our proposed model, GLIMS, we perform 3D segmentation experiments on two datasets, namely Brain Tumor Segmentation (BraTS) 2021 \cite{bakas2017advancing} and Beyond the Cranial Vault (BTCV) Abdomen Dataset \cite{gibson_2018_1169361}. Each task was conducted in 5-fold cross-validation setting. The details of the datasets are provided in the following sections.

\subsubsection{Brain Tumor Segmentation (BraTS)}

The BraTS dataset \cite{bakas2017advancing} comprises 1,251 multi-institutional 3D brain MRI scans in four modalities - T1, T1c, T2, and FLAIR - with corresponding glioma segmented masks, which have four labels: necrotic tumor core (NCR), peritumoral edematous tissue (ED), enhancing tumor (ET), and the background. Each modality's cross-sectional images are registered, and the skull is removed. The slices exhibit a high-resolution isotropic voxel size of $1 \times 1 \times 1$ $mm^3$, and each MRI scan is of dimensions $240 \times 240 \times 155$ in height, width, and depth. To align with existing literature, the provided mask labels are transformed into new label groups: Whole Tumor (WT) (NCR + ED + ET), Tumor Core (TC) (NCR + ET), and Enhancing Tumor (ET).

\subsubsection{Beyond the Cranial Vault (BTCV) Abdomen Dataset}

The BTCV \cite{gibson_2018_1169361} dataset contains CT scans of 30 subjects with annotations for 13 organs: \textit{spleen, right kidney, left kidney, gallbladder, esophagus, liver, stomach, aorta, inferior vena cava, portal vein and splenic vein, pancreas, right adrenal gland, left adrenal gland.} The CT scans have resolutions varying between $512 \times 512 \times 85\sim198$, with a voxel spacing of $0.54 \times 0.98 \times [2.5\sim5.0]$ $mm^3$. 

\begin{table*}[t!]
\centering
\caption{Quantitative results of the BTCV dataset in the mean 5-fold cross-validation Dice Score results for five recent state-of-the-art models. The \textbf{bold} scores indicate the best results, and the \underline{underlined} ones represent the second-best-performing models. Spl: \textit{Spleen}, RKid: \textit{Right Kidney}, LKid: \textit{Left Kidney}, Gall: \textit{Gallbladder}, Eso: \textit{Esophagus}, Liv: \textit{Liver}, Sto: \textit{Stomach}, Aor: \textit{Aorta}, IVC: \textit{Inferior Vena Cava}, Veins: \textit{Portal and Splenic Veins}, Pan: \textit{Pancreas}, AG: \textit{Adrenal Gland}.}
\label{tab:BTCV}
\resizebox{\textwidth}{!}{%
\begin{tabular}{cccccccccccccc}
\toprule
Methods  &
  mDSC ($\%$)  $\uparrow$ &
  Spl  &
  Rkid  &
  Lkid  &
  Gall  &
  Eso  &
  Liv  &
  Sto  &
  Aor  &
  IVC  &
  Veins  &
  Pan  &
  AG  \\ \hline
  UNETR \cite{hatamizadeh2022unetr}    & 75.92 & 89.46 & 90.29 & 86.40 & 64.77       & 64.11          & 93.88 & 62.16       & 82.90 & 70.33 & 62.99          & 80.12 & 63.62 \\
  nnU-Net \cite{isensee2021nnu}  & 81.85 & 94.22 & 89.43 & 90.06 & \underline{73.83} & 72.65          & 94.84 & 82.38       & 87.73 & 78.17 & \textbf{78.31} & 78.93 & 61.60 \\
TransBTS \cite{wang2021transbts} & 82.12 & 94.59 & 89.23 & 90.47 & 68.50       & \textbf{75.59} & 96.14 & 83.72       & 88.85 & 82.28 & 74.25          & 75.12 & 66.74 \\
nnFormer \cite{zhou2021nnformer} & 82.30 & 94.51 & 88.49 & 93.39 & 65.51       & \underline{74.49}    & 96.10 & \underline{83.83} & 88.91 & 80.58 & \underline{75.94}    & 77.71 & 68.19 \\
Swin UNETR \cite{hatamizadeh2021swin} &
  \underline{83.57} &
  \underline{95.57} &
  \underline{93.90} &
  \underline{93.47} &
  70.15 &
  73.47 &
  \underline{96.30} &
  82.20 &
  \underline{89.39} &
  \underline{84.77} &
  74.70 &
  \underline{80.33} &
  \underline{68.54} \\ \hline
\textbf{GLIMS} &
  \textbf{84.50} &
  \textbf{95.81} &
  \textbf{94.39} &
  \textbf{93.93} &
  \textbf{75.64} &
  74.13 &
  \textbf{96.82} &
  \textbf{84.77} &
  \textbf{89.90} &
  \textbf{85.66} &
  73.22 &
  \textbf{80.90} &
  \textbf{68.83} \\ \bottomrule
\end{tabular}%
}
\end{table*}

\begin{table}[]
  \centering
  \caption{Quantitative results of the BraTS2021 dataset in the mean 5-fold cross-validation Dice Score results for five recent state-of-the-art models. The \textbf{bold} scores indicate the best results, and the \underline{underlined} ones represent the second-best-performing models. ET: \textit{Enhancing Tumor}, TC: \textit{Tumor Core}, WT: \textit{Whole Tumor}.}
    \begin{tabular}{ccccc}
    \toprule
    Methods & ET      & TC      & WT      & mDSC (\%) $\uparrow$ \\
    \midrule
    UNETR \cite{hatamizadeh2022unetr}   & 79.78   & 81.57   & 87.66   & 83.00 \\
    TransBTS \cite{wang2021transbts} & 86.80   & 89.80   & 91.10   & 89.23 \\
    SegResNet \cite{myronenko20193d} & 88.30   & 91.30   & 92.70   & 90.77 \\
    nnU-Net \cite{isensee2021nnu} & 88.35   & 91.53   & 92.84   & 90.91 \\
    Swin UNETR \cite{hatamizadeh2021swin} & \underline{89.10}   & \underline{91.70}   & \underline{93.30}   & \underline{91.37} \\
    \midrule
    \textbf{GLIMS}   & \textbf{90.13} & \textbf{92.93} & \textbf{93.37} & \textbf{92.14} \\
    \bottomrule
    \end{tabular}%
  \label{tab:BraTS}%
\end{table}%

\subsection{Evaluation Metrics}

The comparison between the proposed approach and the literature is conducted with two metrics: Dice Similarity Coefficient (DSC) and 95\% Hausdorff Distance (HD95). The DSC assesses the overlap between the predicted volumetric mask and its corresponding volumetric ground truth, as defined in Equation \ref{eq:Dice}, where \textit{Y} represents the predicted mask, and \textit{P} is the ground truth.

\begin{equation}
    DSC(Y, P) = \frac{2 \times |Y \cap P|}{|Y| \cup |P|}
    \label{eq:Dice}
\end{equation}

HD95 measures the 95th percentile of distances from points on the boundaries of the prediction to the nearest ground truth mask, as defined in Equation \ref{eq:HD95}, where $\textbf{d}_{YP}$ represents the maximum 95th percentile distance between predicted voxels and the ground truth, while $\textbf{d}_{PY}$ signifies the maximum 95th percentile distance between the ground truth and the predicted voxels.

\begin{equation}
    \text{HD}_{95}(A, B) = \max\{\textbf{d}_{YP}, \textbf{d}_{PY}\}
    \label{eq:HD95}
\end{equation}

\subsection{Implementation Details}

The GLIMS model was developed using the PyTorch framework v2.0.1 and implemented within the MONAI library v1.2.0. In order to ensure a fair comparison, identical pre-processing strategies and augmentation techniques were employed across models. The models were trained utilizing the AdamW optimizer with a cosine annealing scheduler. The models were trained up to 1000 epochs. The chosen batch size was two, and a sliding window approach with a 0.8 inference overlap was applied using $96 \times 96 \times 96$ patch size on an NVIDIA GTX 3090 GPU with 24GB of memory. The initial learning rate was set to 1e$^{-3}$ and a weight decay of 1e$^{-5}$. We optimized the models by utilizing the average dice loss and cross-entropy loss. The evaluation states with the highest DSC scores are saved and the comparisons were conducted based on these best states.

\subsection{Comparison with State-of-the-Art Methods}

The following sections present comparative experimental results of GLIMS on BraTS2021 and BTCV datasets with well-known state-of-the-art models.

\subsubsection{Glioblastoma Segmentation on BraTS}

For the BraTS2021 dataset, the average of 5-fold cross-validation segmentation results are presented in Table \ref{tab:BraTS}. GLIMS achieves an average DSC of 90,13\%, 92,93\%, and 93,37\% for ET, TC, and WT classes. We compared our results with five methods that can be grouped as CNN-based models (nnU-Net, SegResNet) and CNN-Transformer-based hybrid models (UNETR, TransBTS, Swin UNETR). Compared with UNETR, our work significantly increased the performance by 9,14\% and improved the result by 0.77\% compared to Swin UNETR. Furthermore, GLIMS outperformed the segmentation performance of the fully-optimized CNN-based method nnU-Net by 1,23\%. 

Regarding the qualitative results, the mask predictions of the models for three cases are shown in Figure \ref {fig:BraTS_Seg}. In the first row, the nnU-Net and UNETR misclassified tumor core and edematous regions, whereas GLIMS could determine the edges precisely through the embedded multi-scale feature aggregator modules, assisting in the short-range modeling. Moreover, while Swin UNETR achieved a result close to GLIMS for the chosen slice, it misclassified the tumor core within the enhancing tumor region. This discrepancy is attributed to Swin UNETR's strong performance in global feature extraction but limited capability in short-range modeling. In the second row, all compared models incorrectly classify the edema region into the tumor core. In contrast, GLIMS demonstrates a more precise distinction between these regions, showcasing its effectiveness in local representation. The results demonstrate that GLIMS surpasses the performance of compared models by accurately predicting both uniform and non-uniform regions. This indicates its capability to model short- and long-range dependencies effectively. In the third row, GLIMS exhibits precise segmentation performance by accurately delineating the boundaries of heterogeneous tumor core regions across multiple sub-regions. This ability is attributed to its capacity for local-global feature fusion and guided masking, emphasizing fine-grained details, which allows for the effective representation of large classes such as edema and enhancing tumors. In the final row, GLIMS demonstrated accurate prediction of region boundaries and consistent predictions of large regions by maintaining the exact boundaries between the enhancing tumor and the peritumoral edema, without introducing any misprediction of tumor core regions, as compared to the other models that were compared.

Furthermore, the overall segmentation performance is visualized in Figure \ref{fig:BraTS_3D} as a 3D mesh in three cases. It can be seen that the predicted 3D mesh is closely matched with the ground truth structure of the tumor in three regions. This characteristic makes GLIMS an effective tool for pre-operative planning or evaluating the prognosis of the disease by accurately capturing the fine-grained segmentation of the entire structure. 

\subsubsection{Multi-Organ Segmentation on BTCV}

GLIMS demonstrates advanced segmentation performance in terms of the average 5-fold cross-validation DSC, achieving the highest score of 84.50\%, in comparison to five methods categorized as CNN-based models (nnU-Net, SegResNet) and CNN-Transformer-based hybrid models (UNETR, TransBTS, Swin UNETR). In Table \ref{tab:BTCV}, the individual DSC scores of twelve classes can be found, where our model outperformed the compared models in ten classes and improved the baseline, TransBTS, by 2.38\% DSC on average. 

\begin{figure} [t]
    \centering
    \includegraphics[width=\linewidth]{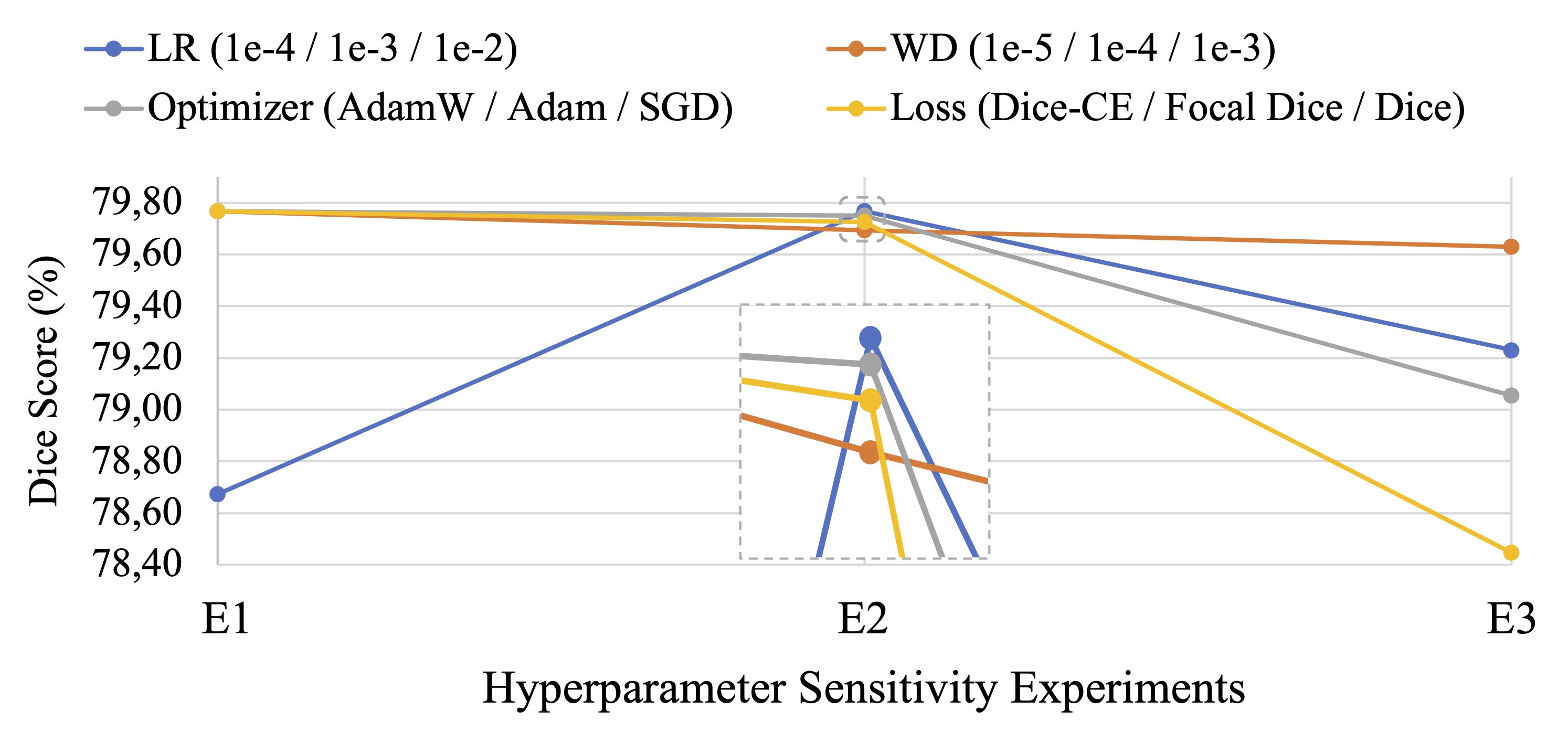}
    \caption{The segmentation performance differences of GLIMS in different hyperparameter settings on the BraTS 2021 dataset. The experiments were performed for 100 epochs.}
    \label{fig:Hyper}
\end{figure}

Quantitatively, the models demonstrate low performance on the right adrenal gland (AG) due to its characteristics of blurry edges and small structures. In addressing this challenge, our model achieved the highest score, surpassing other methods by at least 7.23\%. Similarly, for another small organ, the Gallbladder (Gall), our model achieved an improvement of at least 10.87\% in terms of DSC. This improvement is attributed to the effectiveness of the multi-scale feature aggregator modules in better-capturing features of the small organs before compromising details in lower resolutions. Additionally, GLIMS maintained better performance for larger organs, such as the liver (Liv) and the stomach (Sto), owing to its efficient fusion of local and global features via the transformer bottleneck.

Based on the qualitative results on multi-organ segmentation in Figure \ref{fig:BTCV_Seg}, it is observed that in the first row, Swin UNETR struggled to accurately segment the pancreas region, resulting in discontinuities in the segmentation. Additionally, UNETR under-segmented the pancreas and stomach regions. This suggests that the complex structure of the model demands more data and a longer training process to achieve accurate segmentation results. In contrast, GLIMS demonstrates correct segmentation of the regions compared to the ground truth masks, indicating its potential to be trained with fewer epochs and less data due to its lightweight and robust design. In the second sample, both Swin UNETR and UNETR under-segmented the pancreas region, while all models overly segmented the left kidney compared to the ground truth. Contrarily, GLIMS exhibited the best performance, leveraging its high capability in local and global feature fusion to assist in the segmentation of large regions like the pancreas.
Furthermore, it prevented the over-segmentation of the left kidney by precisely determining the borders with multi-scale feature aggregators. In the third row, GLIMS demonstrated superior performance in determining edges and fully segmenting large areas owing to its hybrid architecture and attention localization during the prediction phase. Conversely, the other methods struggled to properly detect the veins region in the liver, leading to an under-segmentation problem. In the final row of the qualitative results, it was noticed that GLIMS was unable to segment the Stomach region entirely. This is believed to be due to the presence of air inside the organ, which could be contained in multiple organs. It was observed that GLIMS was able to successfully segment the left adrenal region, while the other methods tended to move into outer areas by over-segmenting. This superior performance of our model can be attributed to the parallel utilization of multi-scale feature extractors, which strongly enhances the local feature representation compared to other hybrid models.

Moreover, the overall segmentation performance is demonstrated in Figure \ref{fig:BTCV_3D} as a 3D mesh in four cases. It is clear that the predictions closely resemble the ground truth structure of the organs. The results highlight the notable capability of GLIMS in preserving fine-grained details at borders and accurately determining organ regions, which positions GLIMS as a promising model for 3D organ segmentation tasks, offering the advantages of ease of training and higher accuracy in segmentation.

\begin{figure} [t]
    \centering
    \includegraphics[width=\linewidth]{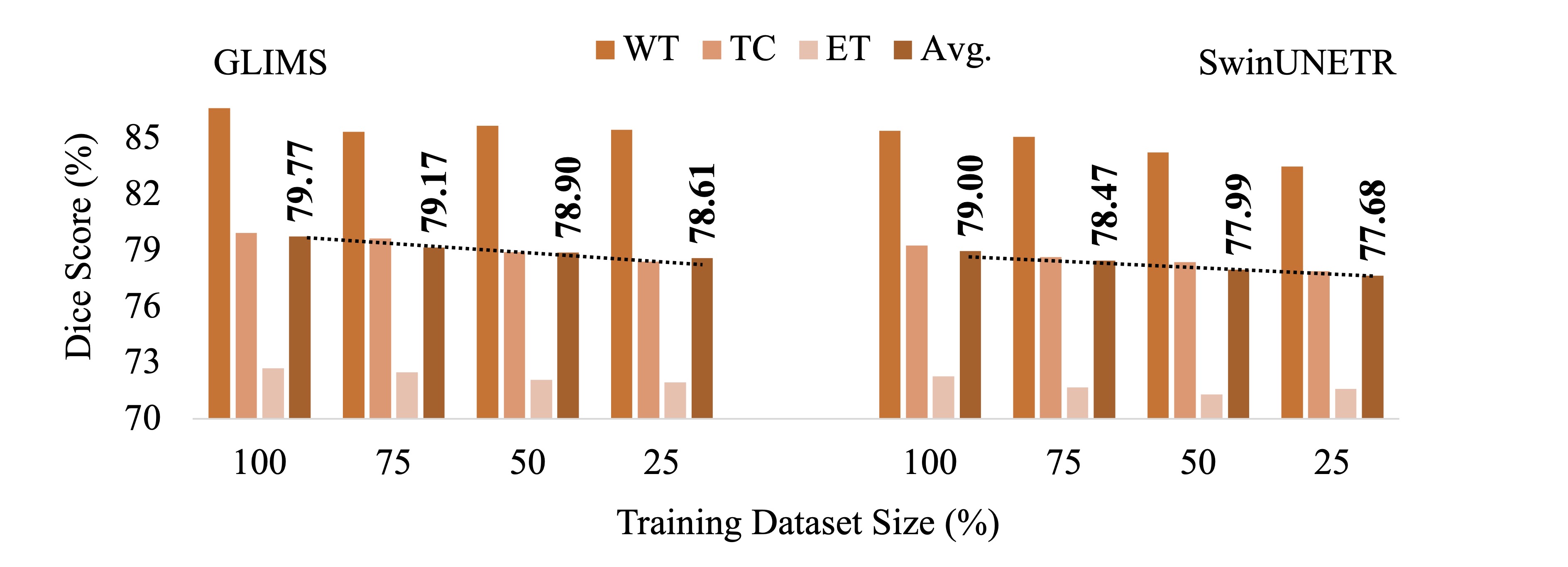}
    \caption{The validation set performances of GLIMS (left) and SwinUNETR (right) on varying training set sizes in the BraTS2021 task. The results are given in WT, TC, ET regions, and the average validation dice score. The experiments were performed for 100 epochs.}
    \label{fig:DatasetSize}
\end{figure}

\subsection{Ablation Studies}

To validate the architectural design of GLIMS, we conducted ablation studies to assess segmentation performance differences on the BraTS2021 dataset, as shown in Table \ref{tab:Ablation}. We focus on evaluating the benefits of the proposed modules -- DMSF, DMSU, CSAB, and transformer bottleneck -- while also considering the impact of the deep-supervision method. The experiments were performed under the same training, augmentation, and hyperparameter settings.

\subsubsection{Evaluation of Hyperparameter Sensitivity}

\begin{figure} [t]
    \centering
    \includegraphics[width=\linewidth]{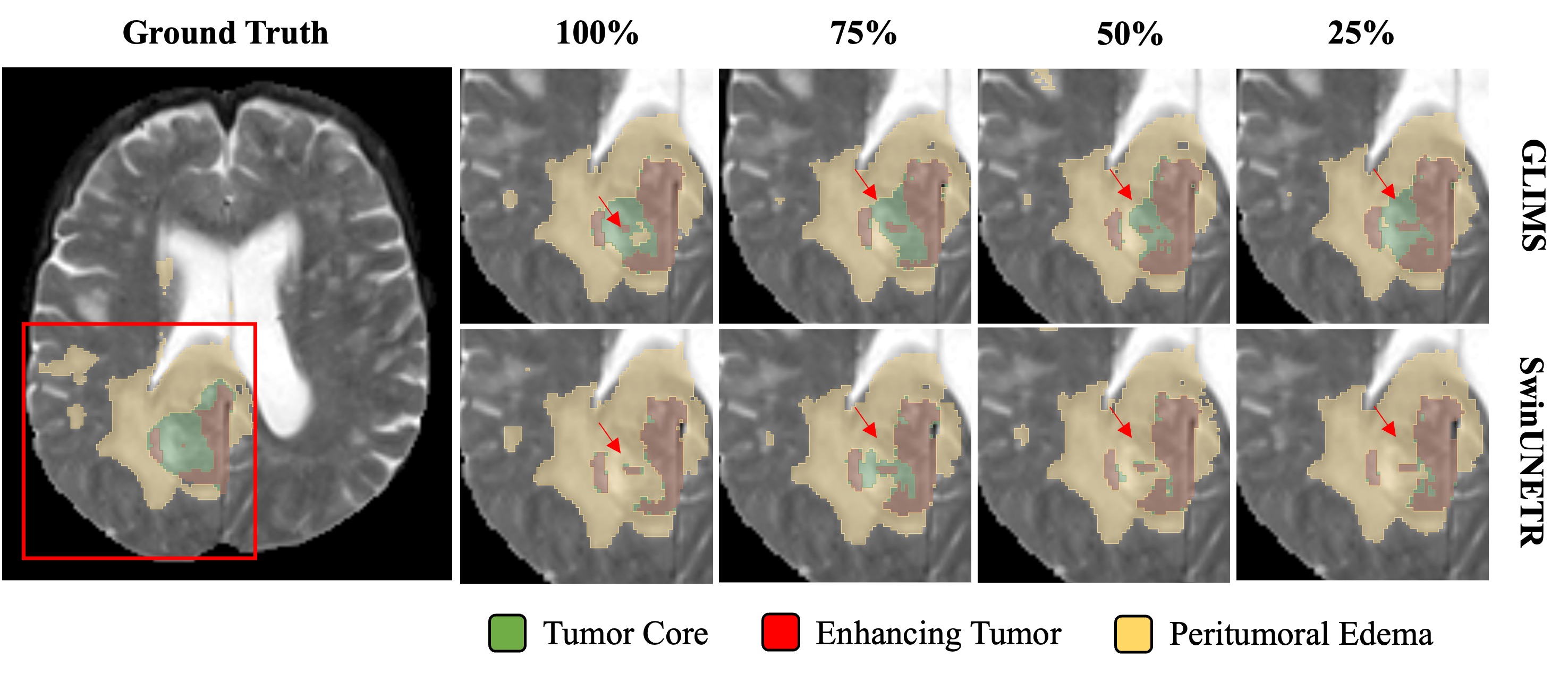}
    \caption{The qualitative results of GLIMS (top) and SwinUNETR (bottom) on the validation set in different training set size selections. Each column represents a decreasing dataset portion from left to right and predictions are made for a single patient on the same MRI slice.}
    \label{fig:DatasetSizeImage}
\end{figure}

To determine how sensitive GLIMS is to changes in its hyperparameters, we conducted a series of experiments on five specific hyperparameters: learning rate, weight decay constant, optimizer, loss function, and training data size. These experiments, as illustrated in Figure \ref{fig:Hyper} and Figure \ref{fig:DatasetSize}, were designed to evaluate the variance in model behavior over a training duration of 100 epochs.

We investigated the impact of various learning rate initialization values, including 1e$^{-2}$, 1e$^{-3}$, and 1e$^{-4}$. During the training of the model, we used the cosine annealing scheduler to adjust the learning rate based on current and total epoch numbers. We discovered that maintaining a higher learning rate of 1e$^{-3}$ resulted in better performance during training. We also analyzed the effect of the weight decay parameter on the model's segmentation performance and found that modifying it did not have a significant impact since the model did not experience overfitting. Therefore, the regularization term did not enhance or diminish its performance. However, we observed notable differences in the choice of optimizers and loss functions. The performance gap between the SGD and AdamW optimizers was 0.72\%, with the Adam and AdamW optimizers performing the best. In the SGD scenario, the momentum parameter requires further tuning to achieve better optimization. Finally, we conducted experiments on three loss functions and found that the Dice-Cross Entropy and Focal-Dice Loss functions produced the highest results, with a 1.32\% decrease in the experiment with Dice Loss. We can conclude that although our model is not sensitive to the learning rate and weight decay coefficient, depending on the task, different optimizer and loss function selections should be tested.

Furthermore, we tested the GLIMS and SwinUNETR models on the BraTS 2021 dataset with training dataset sizes of 25\%, 50\%, 75\%, and 100\%. As the training set size decreased, GLIMS experienced a decrease in mean Dice Score of 1.16\% between full and 25\% datasets, while the SwinUNETR model's performance reduced by 1.32\%, showing the robustness of GLIMS in limited dataset scenarios. The results of the experiment are presented visually in Figure \ref{fig:DatasetSizeImage}. The analysis revealed that in all experiments, GLIMS accurately identified the tumor core region in the predictions. On the other hand, SwinUNETR tends to over-segment the region of peritumoral edema. While SwinUNETR is capable of representing global features, it struggles to identify small details. Conversely, GLIMS features multi-scale extraction modules to address this limitation.

\subsubsection{Evaluation of Deep Supervision}

To assess the impact of deep supervision, we conducted a single-fold validation by comparing a Swin Transformer-based encoder and decoder structured architecture, similar to Swin-UNet \cite{cao2022swin}, against the same model with deep supervision. The results demonstrate that incorporating deep supervision in the loss function improved segmentation performance to a DSC of 90.64 and an HD95 of 7.59 mm, representing a change of +0.5\% and -0.62 mm, respectively. This improvement can be attributed to the benefits of deep supervision in addressing the vanishing gradient problem, facilitating faster convergence, enhancing feature learning at multiple levels as a regularization technique to prevent overfitting, and facilitating better error localization at lower layers. Therefore, contributing to enhanced learning and generalization on complex tasks.

\begin{figure} [t]
    \centering
    \includegraphics[width=\linewidth]{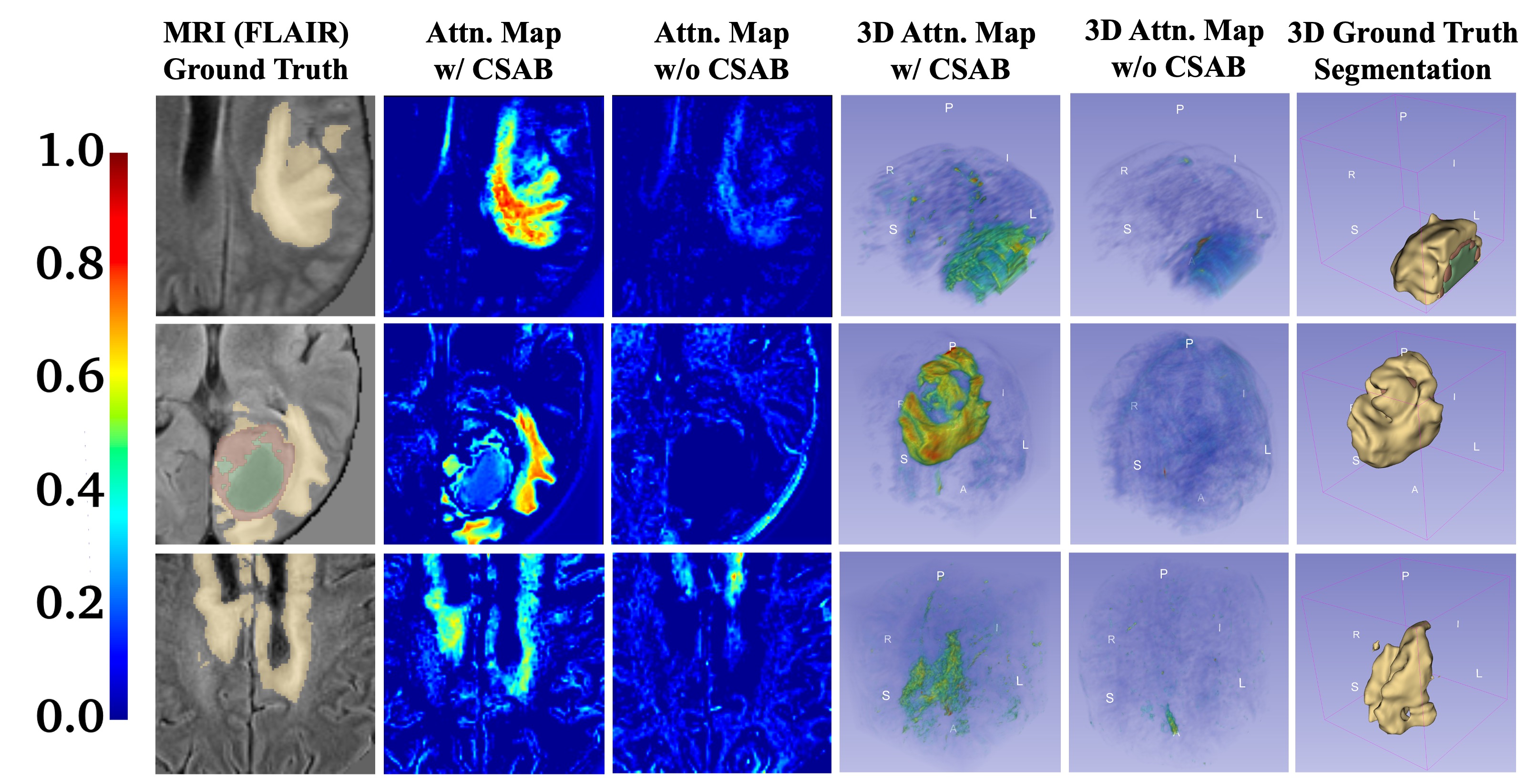}
    \caption{The Grad-CAM \cite{selvaraju2017grad} visualization of the attention maps extracted with the CSAB module in the first level of the skip connection. The figure shows the MRI slice, the attention map with and without the CSAB module, the 3D attention map with and without CSAB, and the 3D ground truth segmentation map, respectively. The color scale indicates the level of attention.}
    \label{fig:Attention}
\end{figure}

\begin{table*}[t]
\centering
\caption{The ablation studies of GLIMS on the BraTS2021 dataset on a single validation set. The \textbf{bold} scores represent the best-performing results. TE: \textit{Transformer Encoder}, TD: \textit{Transformer Decoder}, DS: \textit{Deep Supervision}, MS: \textit{Multi-Scale Feature Aggregation}, AG: \textit{Attention Guidance}, TB: \textit{Transformer Bottleneck}.}
\label{tab:Ablation}
\resizebox{\textwidth}{!}{%
\begin{tabular}{llllllllccccc}
\toprule
{ }  &                                      & \multicolumn{6}{c}{Methods} &       &       &       & \multicolumn{2}{c}{Average} \\ \cline{3-8} \cline{12-13}
\multirow{-2}{*}{{ \#}} &
  \multirow{-2}{*}{Models} &
  TE &
  TD &
  DS &
  MS &
  AG &
  TB &
  \multirow{-2}{*}{ET} &
  \multirow{-2}{*}{TC} &
  \multirow{-2}{*}{WT} &
  DSC ($\%$) $\uparrow$ &
  HD95 (mm) $\downarrow$ \\ \hline
{0} & Baseline (Conv. Enc. \& Dec.)        &     &    &    &    &    &   & 87.70 & 90.16 & 91.29 & 89.72         & 8.92        \\
{1} & Swin Trans. Enc + Conv. Dec.        & \checkmark   &    &    &    &    &   & 87.86 & 90.29 & 92.14 & 90.10         & 8.68        \\
{2} & Swin Trans. Enc. \& Dec.             & \checkmark   & \checkmark  &    &    &    &   & 88.19 & 90.35 & 91.88 & 90.14         & 8.21        \\
{3} & Swin Trans. Enc. \& Dec. + Deep Sup. & \checkmark   & \checkmark  & \checkmark  &    &    &   & 88.49 & 91.42 & 92.00 & 90.64         & 7.59        \\
{4} & Deep Sup. + DMSF  + DMSU             &     &    & \checkmark  & \checkmark  &    &   & 89.48 & 91.31 & 92.32 & 91.04         & 5.54        \\
{5} & Deep Sup. + DMSF  + DMSU   + CSAB    &     &    & \checkmark  & \checkmark  & \checkmark  &   & 89.66 & 91.68 & 92.88 & 91.41         & 5.44        \\
{6} &
  Deep Sup. + DMSF  + DMSU   + CSAB + Trans. Bott. &
   &
   &
  \checkmark &
  \checkmark &
  \checkmark &
  \checkmark &
  \textbf{90.13} &
  \textbf{92.93} &
  \textbf{93.37} &
  \textbf{92.14} &
  \textbf{4.34} \\ \bottomrule
\end{tabular}%
}
\end{table*}

\subsubsection{Evaluation of Multi-Scale Feature Aggregators}

The impact of the feature-extracting modules DMSF and DMSU was assessed by comparing the deeply supervised transformer-based network with a model incorporating an encoder-decoder branch featuring multi-scale feature aggregating layers, in which the DACB modules are comprised. The results show that replacing the encoder and decoder branches with the proposed DACB blocks enhances the DSC score to 91.04\% and HD95 to 5.54 mm, indicating +0.4\% and -2.05 mm improvements, respectively. This improvement can be attributed to enhanced local and global feature extraction at higher levels, leading to more precise segmentations on boundaries for small objects and consistent segmentation for large regions. On the other hand, as illustrated in Figure \ref{fig:BraTS_Seg} and Figure \ref{fig:BTCV_Seg}, architectures with transformer-based encoders, such as Swin UNETR and UNETR, performed well on large region segmentations due to their long-range feature representation abilities, but resulted in less accurate boundaries. Therefore, it is evident that incorporating multi-scale encoder and decoder branches can be advantageous for segmentation tasks involving heterogeneous regions to learn the structure of the regions properly. We further conducted ablation studies to investigate the effect of varying dilation rates. Our findings indicated that incorporating dilation rates across different scales led to a gradual improvement in performance, as shown in Table \ref{tab:Ablation2}

\begin{table}[t]
  \centering
  \caption{The ablation studies on the depth of the transformer layers and the dilation rate combinations of the proposed DACB module. The \textbf{bold} parameters represent the GLIMS's settings. $S_p$: The patch size as a voxel. $S_e$: The embedding dimension in the bottleneck. $T_1$, $T_2$ and $T_b$: The depth of the first, second, and bottleneck transformer layers, respectively. Dilation refers to the individual or combined dilation rates in the DACB module.}
  \resizebox{\columnwidth}{!}{
    \begin{tabular}{ccccccccc}
    \toprule
    \multirow{2}[3]{*}{Models} & \multicolumn{2}{c}{Dimension} & \multicolumn{3}{c}{Depth}   & \multicolumn{1}{c}{\multirow{2}[3]{*}{Dilation}} & \multicolumn{1}{r}{\multirow{2}[3]{*}{mDSC (\%) $\uparrow$}} & \multirow{2}[3]{*}{Params (M)} \\
\cmidrule{2-6}            & $S_p$      & $S_e$      & \multicolumn{1}{l}{$T_1$} & \multicolumn{1}{l}{$T_2$} & \multicolumn{1}{l}{$T_b$} &         &         &  \\
\midrule
    \textbf{GLIMS} & \textbf{96} & \textbf{768} & \textbf{2} & \textbf{6} & \textbf{2} & \textbf{1/2/3} & \textbf{92.14} & \textbf{47.16} \\
    \midrule
    A       & 96      & 768     & 2       & 6       & 2       & \textcolor[rgb]{ 1,  0,  0}{1/2} & 91.58   & 46.73 \\
    B       & 96      & 768     & 2       & 6       & 2       & \textcolor[rgb]{ 1,  0,  0}{1} & 91.25   & 46.29 \\
    \midrule
    C       & 96      & 768     & \textcolor[rgb]{ 1,  0,  0}{-} & 6       & 2       & 1/2/3   & 91.17   & 44.67 \\
    D       & 96      & 768     & \textcolor[rgb]{ 1,  0,  0}{-} & \textcolor[rgb]{ 1,  0,  0}{-} & 2       & 1/2/3   & 90.27   & 27.74 \\
    \midrule
    E       & 96      & 768     & 2       & \textcolor[rgb]{ 1,  0,  0}{4} & 2       & 1/2/3   & 91.73   & 39.96 \\
    F       & 96      & 768     & 2       & \textcolor[rgb]{ 1,  0,  0}{2} & 2       & 1/2/3   & 91.26   & 32.76 \\
    G       & 96      & 768     & 2       & 2       & \textcolor[rgb]{ 1,  0,  0}{4} & 1/2/3   & 91.35   & 47.04 \\
    H       & 96      & 768     & 2       & 2       & \textcolor[rgb]{ 1,  0,  0}{6} & 1/2/3   & 91.44   & 61.32 \\
    I       & 96      & 768     & \textcolor[rgb]{ 1,  0,  0}{4} & 2       & \textcolor[rgb]{ 1,  0,  0}{2} & 1/2/3   & 91.36   & 34.59 \\
    J       & 96      & 768     & \textcolor[rgb]{ 1,  0,  0}{6} & 2       & 2       & 1/2/3   & 91.52   & 36.42 \\
    \bottomrule
    \end{tabular}%
    }
  \label{tab:Ablation2}%
\end{table}%

\subsubsection{Evaluation of Attention Guidance}

During the mask generation of GLIMS, the attention signals are extracted from each level of the encoder branch, which is then directed to the corresponding layers of the decoder branch. The objective is to guide the model to emphasize important regions during prediction, thereby avoiding misclassifications of the background in the scans. For this purpose, we implemented the CSAB module to refine the encoded features in parallel flows, addressing both channel and spatial dimensions. Furthermore, the provided attention signal is concatenated with the decoded features and processed for predictions, guiding the model in a learnable structure. The Grad-CAM \cite{selvaraju2017grad} visualized attention maps of three sample MRI scans can be seen in Figure \ref{fig:Attention}. It is clear that as the module is integrated, the tumor region gets more attention, and the unnecessary focus is eliminated; therefore, the model could segment the region more accurately. Compared to the model with DMSF and DMSU modules, the CSAB-integrated model achieved a DSC of 91.41\% and an HD95 of 5.44 mm, corresponding to an improvement of +0.37\% and -0.1 mm, respectively. Therefore, it can be concluded that attention guidance aids the model in localizing the region of interest and eliminating less important regions to generate a precise prediction. Moreover, since the guidance is performed in a learnable manner, the weights assigned to attention and features are optimized by the model during the training process.

\subsubsection{Evaluation of Transformer Bottleneck}

A Swin Transformer-based bottleneck was integrated into GLIMS to leverage local context information fused with global representations for precise region identification. The transformer layer begins from the last two layers of the encoder and continues until the end of the first two layers of the decoder branch, hybridly fusing features with CNN layers. As the MSA layers are computationally expensive regarding trainable parameters, employing a hybrid approach through both the encoder and decoder branches was not deemed suitable for generating an accurate yet lightweight model, especially critical for 3D segmentation tasks. To evaluate the current design, we compared the segmentation performance with fully CSAB-embedded and CNN-based architecture. The results show that the design achieved 92.14\% DSC and 4.34 HD95, improving the previous design by +0.73\% and -1.1 mm, respectively. During our investigation, we examined how different depths of the transformer layers affected our model, as given in Table \ref{tab:Ablation2}. We found that increasing the depth of the bottleneck layer $T_b$ led to a significant increase in the model size, while changes in the initial layers had a lesser impact. Additionally, we discovered that incorporating more transformer layers in $T_2$ resulted in a greater performance increase compared to $T_1$ or $T_b$. Based on this, we concluded that the optimal depth for $T_1$, $T_2$, and $T_b$ was 2/6/2, respectively.

Observing the qualitative results in Figure \ref{fig:BraTS_Seg} and Figure \ref{fig:BTCV_Seg}, it can be noted that the transformer bottleneck contributes to accurately segmenting large regions without defects, while the precise segmentation of borders is facilitated by the local representation fusion obtained from DMSF and DMSU blocks.

\begin{table}[t]
\centering
\caption{Comparison of the number of parameters, FLOPs, inference time (s), and Dice Scores on the BraTS2021 and BTCV datasets among the selected recent CNN and transformer-based architectures.}
\label{tab:Complexity}
\resizebox{\columnwidth}{!}{%
\begin{tabular}{cccccc}
\hline
\multirow{2}{*}{Methods} & \multirow{2}{*}{Params (M)} & \multirow{2}{*}{FLOPs (G)} & \multirow{2}{*}{Inference Time (s)} & \multicolumn{2}{c}{DSC (\%) $\uparrow$} \\
                                &       &        &       & BraTS2021 & BTCV  \\ \hline
UNETR                           & 92.49 & 75.76  & 10.05 & 83.00     & 75.92 \\
TransBTS                        & 32.79 & 330.37 & 25.02 & 89.23     & 82.12 \\
nnU-Net                         & 19.07 & 412.65 & 15.79 & 90.91     & 81.85 \\
nnFormer                        & 150.5 & 213.4  & 30.12 & -         & 82.30 \\
Swin UNETR                      & 62.83 & 384.2  & 21.97 & 91.37     & 83.57 \\ \hline
\textbf{GLIMS} & 47.16 & 72.30  & 36.63 & \textbf{92.14}     & \textbf{84.50} \\ \hline
\end{tabular}%

}
\end{table}

\subsubsection{Evaluation of Computational Complexity}

We present the computational complexities of the evaluated models in Table \ref{tab:Complexity} in terms of trainable parameters, the number of FLOPs, and the average DSC values in two volumetric segmentation tasks on BraTS2021 and BTCV datasets. The number of trainable parameters and FLOPs are computed using a sliding window approach with an input size of $96 \times 96 \times 96$. In contrast to transformer-based architectures, models primarily composed of CNN blocks, such as nnU-Net and TransBTS, exhibit fewer trainable parameters, i.e., 19.07M and 32.79M, respectively. Regarding transformer-based architectures and models incorporating the self-attention mechanism, such as UNETR, nnFormer, and Swin UNETR, they showcase a higher number of trainable parameters, amounting to 92.49M, 150.5M, and 61.98M, respectively, primarily due to the characteristics of the layers. This complexity makes them data-intensive and necessitates longer training iterations for convergence. In contrast, GLIMS features a relatively low number of trainable parameters and floating-point operations. In terms of the inference time, GLIMS takes 36.63 seconds to conduct a thorough segmentation of a 3D scan. Although our model has a reduced number of FLOPs, it includes multi-scale convolutional feature extractors, which results in a slightly longer inference time due to the sliding window nature of the convolution operation. Compared to nnU-Net's fully convolutional architecture, our model shows an increase in inference time of 20.84 seconds and in terms of the transformer-based SwinUNETR model, GLIMS exhibits an increase in the inference time by 14.66 seconds.

The given attributes of GLIMS position it as a promising architecture for volumetric segmentation tasks, as demonstrated by its state-of-the-art results with 92.14\% and 84.50\% DSC values on BraTS2021 and BTCV datasets, respectively. The efficient fusion of depth-wise CNN and transformer layers, coupled with the incorporation of attention-guided segmentation, contributes to GLIMS' high-performance character.

\section{Conclusions}

This study introduces GLIMS, a novel hybrid CNN-transformer-based method for semantic segmentation of volumetric medical images. Our approach leverages multi-scale contextual features extracted from the encoder branch through DACB blocks to enhance the model's representation ability. We exploit the fusion of local and global representations using Swin Transformer-based bottleneck layers and perform attention guidance via CSAB modules on the decoder branch to generate fine-grained 3D segmentation outputs. GLIMS achieved a notable performance in both segmentation ability and model complexity when compared to well-known CNN-based and hybrid methods, particularly on the 3D multi-organ segmentation dataset, BTCV, and the volumetric glioblastoma segmentation task. In conclusion, GLIMS exhibits great potential for exploiting long-short-range relations for anatomical reasoning in medical images. It emerges as a novel lightweight and high-performing volumetric segmentation model for medical image analysis, particularly valuable in scenarios with limited available data. Moreover, the design of GLIMS can be used with any 3D imaging modality without requiring modification, specifically in the architecture. 

One of the major challenges in the field is to enhance the capacity of models to learn through hybrid approaches while simultaneously keeping the data requirement of the models low. This is especially important for scenarios where segmentation performance is essential for a specialized area with limited data. Such tasks would require less data to achieve effective segmentation. Furthermore, the integration of hybrid designs should be carried out in such a way that the model's efficiency remains high. In our future work, we aim to extend GLIMS to 2D medical image segmentation tasks, expanding its usability in the area. Additionally, while the Channel and Spatial-Wise Attention Block (CSAB) demonstrated improvement, we plan to explore hierarchical attention layers that combine multiple levels to gain additional insights for generating attention maps rather than relying on a single level.

\section* {CRediT Authorship Contribution Statement}
\textbf{Ziya Ata Yazıcı}: Conceptualization, Methodology, Validation, Software, Investigation, Writing - Review \& Editing. \textbf{İlkay Öksüz}: Conceptualization, Supervision, Writing - Review \& Editing. \textbf{Hazım Kemal Ekenel}: Conceptualization, Supervision, Writing - Review \& Editing.

\section*{Acknowledgements}

This study has been partially funded by Istanbul Technical University, Department of Computer Engineering and Turkcell via a Research Scholarship grant provided to Ziya Ata Yazıcı.



\bibliographystyle{elsarticle-num} 
\bibliography{bib}





\end{document}